\documentclass[letterpaper, 10pt, conference]{ieeeconf/ieeeconf}    % Use this line for a4 paper

\makeatletter
\let\NAT@parse\undefined
\makeatother

\usepackage[numbers,sectionbib,sort&compress]{natbib}

\usepackage{bm}
\usepackage{gensymb}
\usepackage{graphicx}
\usepackage{amsmath}
\usepackage{amssymb}
\usepackage{algorithm}
\usepackage[noend]{algpseudocode}
\usepackage{subcaption}
\usepackage{amsfonts}
\usepackage{siunitx}
\usepackage{booktabs}
\usepackage{upgreek}
\usepackage[font=small]{caption}
\usepackage[export]{adjustbox}
\usepackage{tikz}
\usepackage{sidecap} \sidecaptionvpos{figure}{c}
\DeclareMathOperator*{\argmax}{\arg\!\max}

\DeclareMathOperator{\Tr}{Tr}

\usepackage{hyperref}
\hypersetup{colorlinks,breaklinks,
linkcolor=[rgb]{0.5,0.,0.},
citecolor=[rgb]{0.000,0.427,0.173},
urlcolor=[rgb]{0.031,0.318,0.612}}

\usepackage[nameinlink]{cleveref}
\usepackage{color}
\definecolor{CommentPink}{rgb}{1,0.2,0.5}

\usepackage{comment}
\newcommand{\appropto}{\mathrel{\vcenter{
  \offinterlineskip\halign{\hfil$##$\cr
    \propto\cr\noalign{\kern2pt}\sim\cr\noalign{\kern-2pt}}}}}

\newcommand\edit[1]{{\color{black}#1}}

\usepackage[printonlyused,withpage,nolist,nohyperlinks]{acronym}
\begin{acronym}

\acro{2D}{two-dimensional}

\acro{3D}{three-dimensional}

\acro{AHRS}{attitude and heading reference system}

\acro{AUV}{autonomous underwater vehicle}

\acro{CPP}{Chinese Postman Problem}

\acro{DoF}{degree of freedom}
\acrodefplural{DoF}[DoFs]{degrees of freedom}

\acro{DVL}{Doppler velocity log}

\acro{FSM}{finite state machine}

\acro{IMU}{inertial measurement unit}

\acro{LBL}{Long Baseline}

\acro{MCM}{mine countermeasures}

\acro{MDP}{Markov decision process}
\acrodefplural{MDP}[MDPs]{Markov decision processes}

\acro{POMDP}{Partially Observable Markov Decision Process}
\acrodefplural{POMDP}[POMDPs]{Partially Observable Markov Decision Processes}

\acro{PRM}{Probabilistic Roadmap}
\acrodefplural{PRM}[PRM]{Probabilistic Roadmaps}

\acro{ROI}{region of interest}
\acrodefplural{ROI}[ROIs]{regions of interest}

\acro{ROS}{Robot Operating System}

\acro{ROV}{remotely operated vehicle}

\acro{RRT}{Rapidly-exploring Random Tree}
\acrodefplural{RRT}[RRTs]{Rapidly-exploring Random Trees}

\acro{SLAM}{Simultaneous Localization and Mapping}

\acro{SSE}{sum of squared errors}

\acro{STOMP}{Stochastic Trajectory Optimization for Motion Planning}

\acro{TRN}{Terrain-Relative Navigation}

\acro{UAV}{unmanned aerial vehicle}

\acro{USBL}{Ultra-Short Baseline}

\acro{IPP}{informative path planning}

\acro{FoV}{field of view}
\acrodefplural{FoV}[FoVs]{fields of view}

\acro{CDF}{cumulative distribution function}

\acro{ML}{maximum likelihood}

\acro{RMSE}{Root Mean Squared Error}
\acro{MLL}{Mean Log Loss}

\acro{GP}{Gaussian Process}
\acrodefplural{GP}[GPs]{Gaussian Processes}

\acro{KF}{Kalman Filter}

\acro{IP}{Interior Point}
\acro{BO}{Bayesian Optimization}

\acro{SE}{squared exponential}

\acro{UI}{uncertain input}

\acro{MCL}{Monte Carlo Localization}
\acro{AMCL}{Adaptive Monte Carlo Localization}

\end{acronym}

\IEEEoverridecommandlockouts                           % This command is only needed if
\title{\LARGE \bf
Informative Path Planning for Active \edit{Field} Mapping \\
under Localization Uncertainty
}
\author{$\text{Marija Popovi\'{c}}$, $\text{Teresa Vidal-Calleja}$, $\text{Jen Jen Chung}$, $\text{Juan Nieto}$, $\text{Roland Siegwart}$
\thanks{M. Popovi\'{c}, J. J. Chung, J. Nieto, and R. Siegwart are with the Autonomous Systems Lab., ETH Z\"{u}rich, Z\"{u}rich, Switzerland.
T. Vidal-Calleja is with the Centre for Autonomous Systems at the Faculty of Engineering and IT, University of Technology Sydney, Australia.
Corresponding author: \texttt{mpopovic@ethz.ch}.}%
}

\begin{document}

\maketitle
\thispagestyle{empty}
\pagestyle{empty}

\begin{abstract}
Information gathering algorithms play a key role
in unlocking the potential of robots for efficient data collection
in a wide range of applications.
However, most existing strategies
neglect the fundamental problem of the robot pose uncertainty,
which is an implicit requirement for creating robust, high-quality maps.
To address this issue,
we introduce an informative planning framework
for active mapping
that explicitly accounts for the pose uncertainty in \textit{both} the mapping and planning tasks.
Our strategy exploits a Gaussian Process (GP) model
to capture a target environmental field
given the uncertainty on its inputs.
%This allows us to maintain robust maps,
%which are used for planning information-rich trajectories in continuous space.
For planning,
we formulate a new utility function
that couples the localization and field mapping objectives in GP-based mapping scenarios
in a principled way, without relying on any manually tuned parameters.
%enabling us to trade-off exploration against exploitation in a principled way.
Extensive simulations show that our approach outperforms existing strategies,
with reductions %of up to $45.1\%$ and $6.3\%$
in mean pose uncertainty and map error.
We also present a proof of concept in an indoor temperature mapping scenario.
\end{abstract}

\section{INTRODUCTION} \label{S:introduction}
Rapid technological advancements are inciting
the use of autonomous mobile robots for exploration and data acquisition.
In many marine~\citep{Hitz2017,Jadidi2019}, terrestrial~\citep{Oliveira2017,Carrillo2018}, and airborne~\citep{Popovic2019,Papachristos2017} applications,
these systems have the ability to bridge the spatiotemporal divides
limiting traditional measurement methods
in a safer and more cost-effective manner~\citep{Manfreda2018}.
However, to fully exploit their potential,
algorithms are required for planning efficient informative paths
in complex environments
under platform-specific constraints.

This paper examines the problem of active mapping using a robot,
where the aim is to recover a continuous 2-D or 3-D field, e.g., of temperature, humidity, etc.,
using measurements collected by an on-board sensor.
In similar setups,
most existing strategies~\citep{Popovic2019,Hitz2017,Hollinger2014}
incorrectly assume perfect pose information,
which is an implicit requirement for building high-quality maps
in initially unknown environments.
Our motivation is to improve upon the robustness and accuracy of field reconstructions
by allowing the robot to adaptively trade-off
between gathering new information (exploration)
and maintaining good localization (exploitation).

Despite recent efforts~\citep{Jadidi2019,Papachristos2017,Girard2004,Mchutchon2011},
propagating both the localization and field map uncertainties
into the planning framework in a principled manner
remains an open challenge.
A major issue arises due to
the different ways in which
the target field and robot pose are modeled.
In particular,
our work considers the task of mapping a field using a \ac{GP}
with the robot pose represented as a multivariate Gaussian distribution.
In this setup,
measures of uncertainty in the field and robot pose,
e.g., entropy-based criteria~\citep{Carrillo2018},
are not directly comparable since they are obtained from their respective model variances
which have different units and scales,
and are therefore difficult to couple
in a single utility function
for multi-objective planning.
%balancing between the localization and mapping objectives.
Heuristic methods,
e.g., based on a linear weighting of uncertainties~\citep{Bourgault2002},
are commonly applied;
however, they require careful manual tuning
and are often scenario-specific.

\begin{figure}[!t]
\centering
 \begin{subfigure}[]{0.24\textwidth}
 \vspace{3mm}
  \includegraphics[width=0.75\columnwidth]{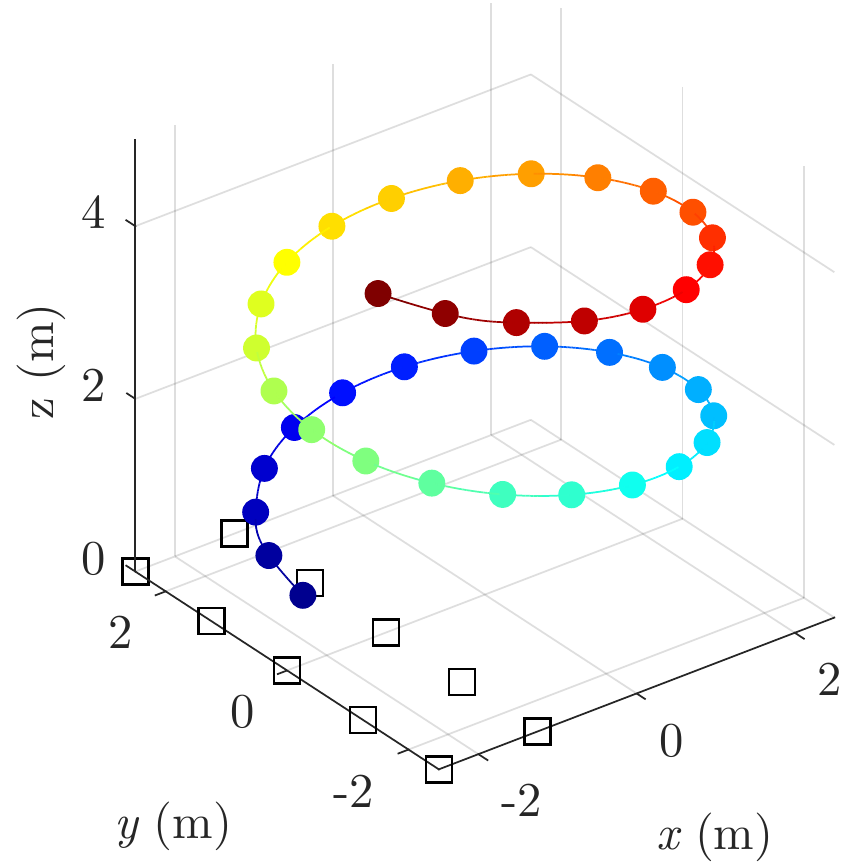}
  \hspace{1mm}
  \raisebox{4mm}{\includegraphics[width=0.17\columnwidth]{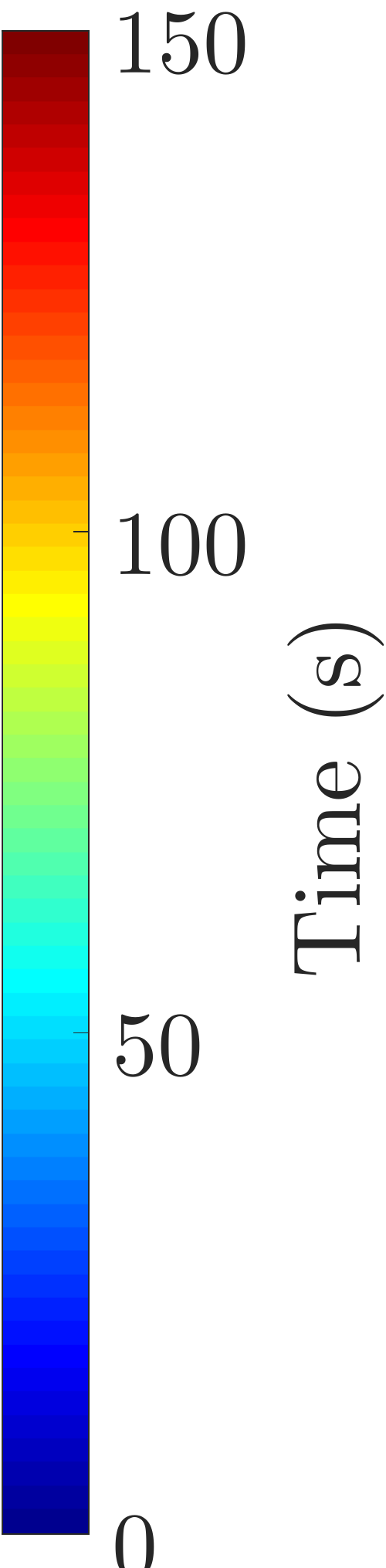}}
  \caption{} \label{SF:teaser_trajectory}
 \end{subfigure}\hfill%
 \begin{subfigure}[]{0.24\textwidth}
  \includegraphics[width=0.75\columnwidth]{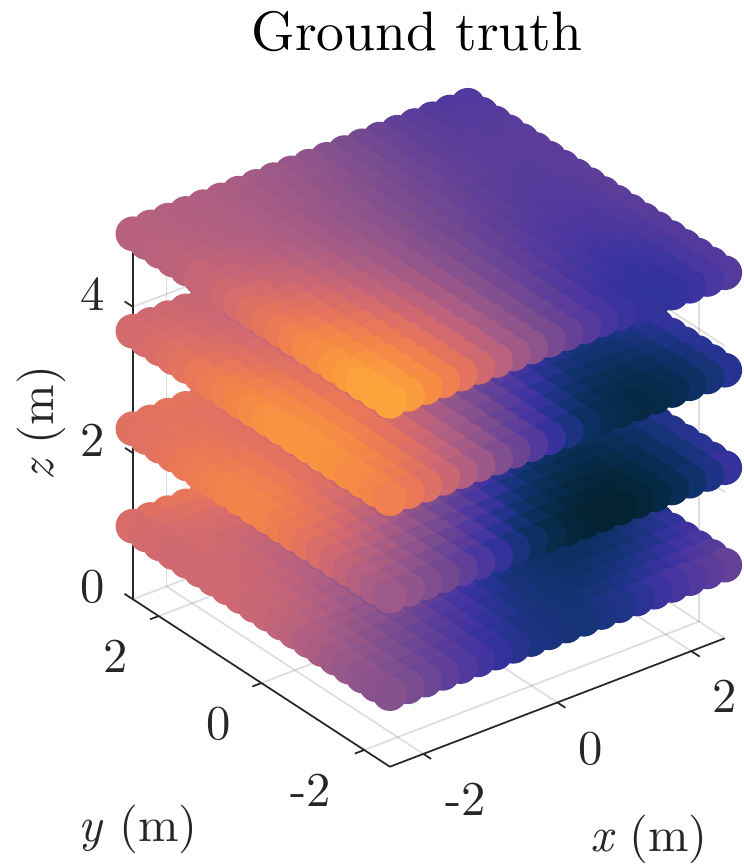}
  \hspace{1mm}
  \raisebox{2mm}{\includegraphics[width=0.17\columnwidth]{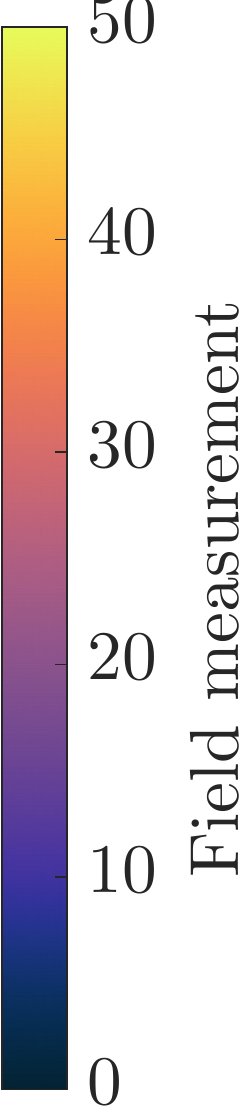}}
  \caption{}
 \end{subfigure}\hfill
   \vspace{1mm}

 \begin{subfigure}[]{0.24\textwidth}
  \includegraphics[width=0.75\columnwidth]{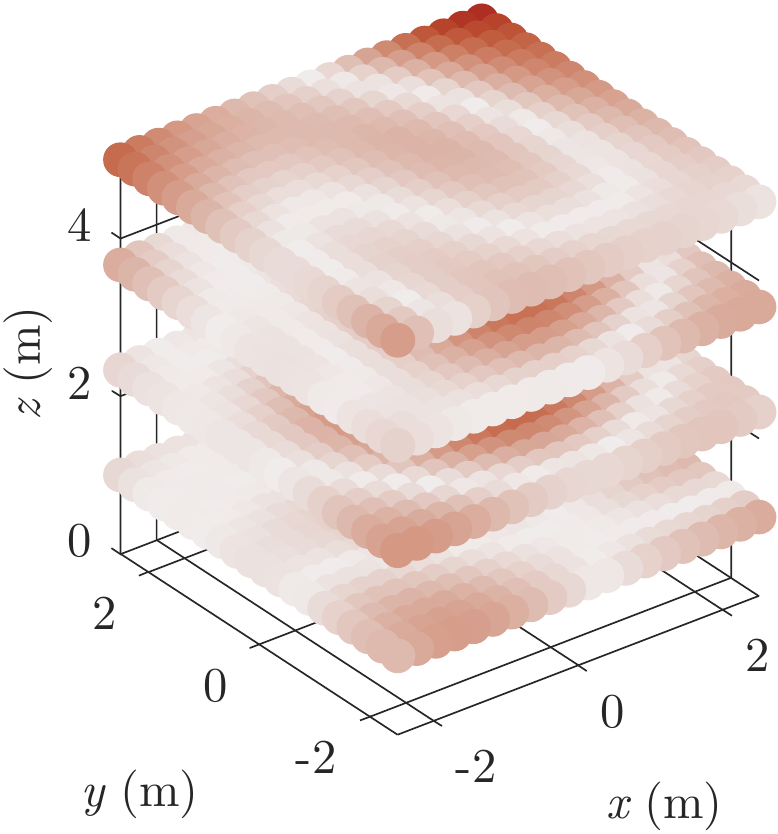}
  \hspace{1mm}
  \raisebox{2mm}{\includegraphics[width=0.17\columnwidth]{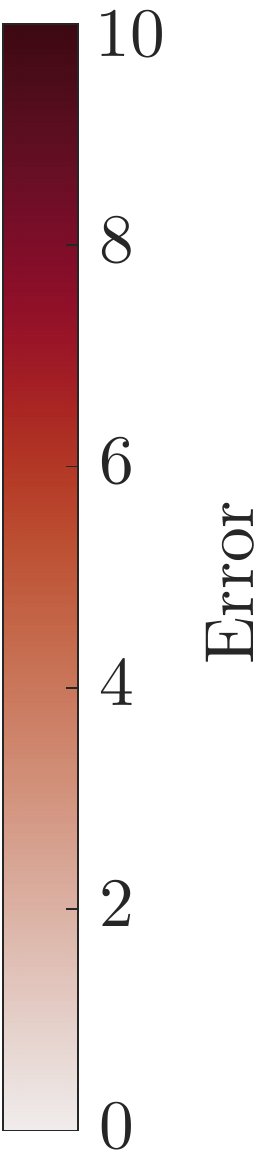}}
  \caption{}
  \end{subfigure}\hfill%
 \begin{subfigure}[]{0.24\textwidth}
  \includegraphics[width=0.75\columnwidth]{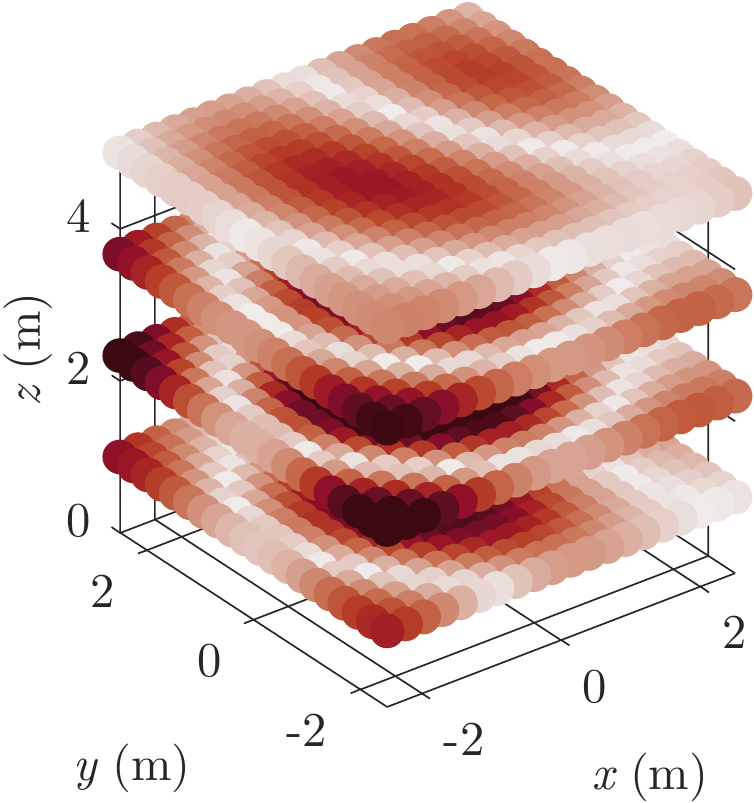}
    \hspace{1mm}
  \raisebox{2mm}{\includegraphics[width=0.17\columnwidth]{images/teaser_error_colorbar.pdf}}
  \caption{}
  \end{subfigure}\hfill%
   \caption{Overview of our proposed active mapping strategy.
   (a) shows a spiral trajectory traveled by a robot.
   The squares indicate point landmarks on the ground used for localization.
   The spheres represent sites where measurements of the ground truth field map in (b) are taken.
   By accounting for pose estimation uncertainty,
   our framework yields an error map (c) with 2.47 times lower total error
   compared to a standard mapping approach (d).
   %Without \acp{UI}, RMSE = 5.09. With \acp{UI}, RMSE = 2.06.
   }\label{F:teaser}
\end{figure}
To address this,
we present an approach that accounts for the pose estimation uncertainty in two places.
First, we consider the additional noise the pose uncertainty induces in the environmental field model;
second, we include it as a shaping factor in the utility function that defines our informative planning task.
Our mapping strategy uses \acp{GP} with \acp{UI}~\citep{Jadidi2019}
to propagate the pose uncertainty into the field model\edit{,
as visualized in \Cref{F:teaser}}.
During a mission,
the map built online is used to plan informative trajectories in continuous space
by optimizing initial solutions obtained by a coarse grid search.
We develop a new utility formulation
for \ac{GP}-based mapping scenarios
that jointly considers the uncertainty of the robot and the field models.
This enables us
to capture the desired exploitation-exploration trade-off in a mathematically sound manner,
without relying on any manually-tuned, environment-dependent parameters.
In summary, the contributions of this work are:
\begin{enumerate}
 \item A utility function based on \ac{GP} field models
  that tightly couples the objectives of robot localization and field mapping
  in active mapping problems.
 \item An informative planning framework that accounts for
 the pose uncertainty in both mapping and the information objective for planning.
 \item Evaluations of the approach in a 3-D graph \ac{SLAM} setup
 and a proof of concept in a temperature mapping scenario.
 %Previously:
 %\item An integrated framework using
 %the proposed utility function for informative planning
 %with maps built online under pose uncertainty,
 %and the evaluation of our approach in a 3-D graph \ac{SLAM} setup.
 %\item Proof of concept in a temperature mapping scenario.
\end{enumerate}
We note that our framework can be used in any scalar field mapping scenario,
e.g., spatial occupancy~\citep{Papachristos2017,Carrillo2018,Valencia2012}, 
signal strength~\citep{Jadidi2019,Hollinger2014},
aerial surveillance~\citep{Manfreda2018,Popovic2019}, etc.,
and with any \ac{SLAM} or localization-only \edit{algorithm
supplying the pose uncertainty.
Specifically, in this work,
we focus on a general case where the spatial field information being collected is decoupled from the localization routine.
More generally,}
the applicability of our utility function extends to other areas of robotics,
such as reinforcement learning~\citep{Whitehead2008}.

\section{RELATED WORK} \label{S:related_work}
Significant recent work has been done on autonomous data gathering strategies
in the context of robotics and related fields.
The discussion in this section focuses on two main research streams:
(1) methods for probabilistic environmental mapping~\citep{Girard2004,Jadidi2019,Papachristos2017,Rasmussen2006},
and (2) algorithms for informative planning~\citep{Popovic2017ICRA,Popovic2017IROS,Popovic2019,Carrillo2018,Valencia2012,Hitz2017}.

\acp{GP} are a popular non-parameteric Bayesian technique for modeling spatio-temporal phenomena~\citep{Rasmussen2006}.
They have been applied in various active sensing scenarios~\citep{Hitz2017,Popovic2019,Popovic2017IROS,Oliveira2017}
to gather data based on correlations and uncertainty in continuous maps.
However,
most of these works assume that the training data for prediction is inherently noise-free,
which may lead to inaccuracies
if measurements are incorporated at wrong locations
and mislead predictive planning algorithms.

Propagating the input uncertainty through dense \ac{GP} models is a computationally challenging task.
To address this,
analytical~\citep{Girard2004} and heteroscedastic approximation methods~\citep{Mchutchon2011,Jadidi2019,Oliveira2017}
have been proposed.
Our work leverages the expected kernel technique of~\citet{Jadidi2019}
by integrating over \acp{UI} with deterministic query points.
Specifically,
we \edit{apply} their approach to more complex planning problems in 3-D scenarios,
and integrate it with a new uncertainty-aware utility function.
This coupling enables robust, tractable mapping under robot pose uncertainty
for online sensing applications.

An active sensing task can be expressed in a \ac{POMDP}~\citep{Kaelbling1998}
as one of decision-making under uncertainty.
In practice,
informative planning algorithms are typically used
to render this problem computationally tractable with dense belief representations.
We broadly distinguish between planning strategies operating in
(a)~discrete~\citep{Valencia2012,Carrillo2018} and (b)~continuous space.
This study focuses on the latter class of methods,
which leverage incremental sampling~\citep{Costante2017,Papachristos2017,Hollinger2014,Bry2011} or splines~\citep{Popovic2019,Hitz2017}
to offer greater scalability compared to discrete approaches.
As in our prior work~\citep{Popovic2017ICRA,Popovic2017IROS,Popovic2019},
we define smooth polynomial robot trajectories~\citep{Richter2013}
and optimize them globally for an information objective
in a finite-horizon manner.

Relatively limited research has been invested in
active sensing scenarios where robot localization is uncertain.
This setup has been tackled in the contexts of
belief-space planning~\citep{Bry2011,Papachristos2017,Costante2017}
and active \ac{SLAM}~\citep{Valencia2012,Bourgault2002},
where the aim is to maintain good localization
as an unknown environment is explored.
In contrast,
and similarly to~\citet{Papachristos2017} and~\citet{Costante2017},
our paper considers a setting where the map building and robot localization problems are decoupled.
An important distinction is that
we aim to reconstruct a continuous field
that is independent of the features used for localization.

An open challenge in this setup is formulating utility functions
to trade off between
robot localization and field mapping in a principled manner.
To address this,
previous approaches have examined heuristic parameter tuning~\citep{Valencia2012},
e.g., using a weighted linear combination of the map and pose uncertanties~\citep{Bourgault2002},
and multi-layer~\citep{Costante2017,Papachristos2017} planning strategies.
In contrast to these methods,
we follow~\citet{Carrillo2018} in using the concept of R\'{e}nyi's entropy
to discount information gain based on predicted localization uncertainty.
Thereby,
our utility function shares the benefit of coupling
the two objectives in a mathematically sound way,
without manual tuning requirements.
The core difference is that our formulation is developed
for a continuous mapping scenario
based on a \ac{GP} field model, instead of an occupancy grid.
Moreover, by mapping with \acp{UI},
we present a unified framework
where the robot localization uncertainty is jointly accounted for
in \emph{both} mapping and planning
to achieve more robust data acquisition.

\section{PROBLEM STATEMENT} \label{S:problem_statement}
The general active mapping problem is formulated as follows.
We seek an optimal trajectory $\psi^*$
in the space of all continuous trajectories $\Psi$
to maximize an information-theoretic measure:
\begin{equation}
\begin{aligned}
  \psi^* ={}& \underset{\psi \in \Psi}{\argmax} \,
  \mathrm{I}(\textsc{measure}(\psi)) \, \textrm{,} \\
  %\frac{\mathrm{I}[\textsc{measure}(\psi)]}{\textsc{cost}(\psi)}\textit{,} \\
  & \text{s.t. } \textsc{cost}(\psi) \leq B \, \textrm{.}
 \label{E:ipp_problem}
\end{aligned}
\end{equation}
The function \textsc{measure(\textperiodcentered)} obtains a finite set of measurements along
trajectory $\psi$,
and \textsc{cost(\textperiodcentered)} provides its associated cost,
which cannot exceed a predefined budget $B$.
The operator $\mathrm{I}($\textperiodcentered$)$ defines the information objective
quantifying the utility of the acquired sensor measurements.
In \Cref{SS:utility_definition},
we propose a utility function
for active mapping in \ac{GP}-based scenarios
that incorporates both the robot localization and field mapping objectives
without any manual parameter tuning requirements.

\section{MAPPING APPROACH} \label{S:mapping_approach}
This section presents our mapping approach as the basis of our framework.
We first describe our method for environmental field modeling using a \ac{GP},
then present a strategy which folds the robot pose uncertainty into the map inference.

\subsection{Gaussian Processes} \label{SS:gaussian_processes}
We use a \ac{GP} to model spatial correlations \edit{of a field} in a probabilistic and non-parametric manner~\citep{Rasmussen2006}.
The target field variable for mapping is assumed to be a continuous function: $f: \mathcal{E} \rightarrow \mathbb{R}$.
%where $\mathcal{E} \subset \mathbb{R}^3$ is the environment where measurements are taken. 
%A Gaussian correlated prior is placed over the function space, which
\edit{Essentially, a \ac{GP} is a generalization of the Gaussian probability distribution, where the stochastic process governs the properties of functions, as opposed to vectors or scalars in the case of the probability distribution. A GP} is fully characterized by the mean function $m(\mathbf{x}) \triangleq \mathop{{}\mathbb{E}}[f(\mathbf{x})]$
and covariance function $k(\mathbf{x},\mathbf{x}') \triangleq \mathop{{}\mathbb{E}}[(f(\mathbf{x})-m(\mathbf{x}))(f(\mathbf{x'})-m(\mathbf{x'})]$
as $f(\mathbf{x}) \sim \mathcal{GP}(m(\mathbf{x}),\, k(\mathbf{x},\mathbf{x}'))$,
where $\mathop{{}\mathbb{E}}[$\textperiodcentered$]$ is the expectation operator
and $\mathbf{x}$ and $\mathbf{x}'$ are input vectors composed of spatial coordinates in the environment.

%Let $\mathbf{x}_i \subset \mathcal{E}$ be a set of $n$ observed input training points in the fixed-size environment with associated target values $y_i$.
\edit{Let $\mathbf{X} \subset \mathcal{E}$ be a set of $n$ observed robot locations $\mathbf{x}_i$ in the fixed-size environment with associated target \edit{field} values $\mathbf{y}$.
%For a set of $n^*$ query \edit{locations} ${\mathbf{x}_*}_i \subset \mathcal{E}$,
For a new set of $n^*$ query \edit{locations} ${\mathbf{X}_*} \subset \mathcal{E}$
we can infer the field Gaussian distribution by conditioning in the observed values}
$\mathbf{f}_* \,|\, \mathbf{X},\, \mathbf{y},\, \mathbf{X}_* \sim \mathcal{N}(\boldsymbol{\upmu},\, \mathbf{P})$
as follows~\citep{Rasmussen2006}:
\begin{align}
 \boldsymbol{\upmu} ={}& m(\mathbf{X}_*) + K(\mathbf{X}_*,\mathbf{X})\,[K(\mathbf{X},\mathbf{X}) + \sigma_n^2\mathbf{I}]^{-1} \times &\notag\\
  &(\mathbf{y} - m(\mathbf{X})) \, \textrm{,} \label{E:gp_mean} &\\
 \mathbf{P} ={}& K(\mathbf{X}_*,\mathbf{X}_*) - K(\mathbf{X}_*,\mathbf{X})[K(\mathbf{X},\mathbf{X}) + \sigma_n^2\mathbf{I}]^{-1} \times &\notag\\
  &K(\mathbf{X}_*,\mathbf{X})^\top \, \textrm{,} \label{E:gp_cov}
\end{align}
where $\sigma_n^2$ is a hyperparameter representing the observation noise variance,
$K(\mathbf{X}_*,\mathbf{X})$ denotes the cross-correlation terms between the query and observed points,
and $K(\mathbf{X},\mathbf{X})$ and $K(\mathbf{X}_*,\mathbf{X}_*)$ are the joint covariance matrices for the observed and query points, respectively.
Note that $K($\textperiodcentered$,\,$\textperiodcentered$)$ corresponds to the covariance function $k($\textperiodcentered$,\,$\textperiodcentered$)$ for only one element.

\edit{Note that $k$ encodes our assumptions about the field we are interested in modeling. In this work, we assume that the mapped environmental phenomena is smooth and isotropic; thus common choices of covariance functions are the \ac{SE} and Mat\'ern family~\citep{Rasmussen2006}.}

%To describe environmental phenomena,
%we propose using a constant mean function $m(\textbf{x}) = m_{gp}$
%and the isotropic \ac{SE} kernel common in geostatistical analysis.
%It is defined as~\citep{Rasmussen2006}:
%%
%\begin{equation}
%   k_{SE}(\mathbf{x},\mathbf{x}') =
%   %\sigma^2_f \exp{(-(\textbf{x}-\textbf{x}^*)^T L (\textbf{x}-\textbf{x}^*))} \textit{,}
%   \sigma^2_f \, \exp{\left(-\frac{d^2}{2l^2}\right)} \, \textrm{,} \label{E:se_kernel}
%\end{equation}
% %
%where $d$ is the Euclidean distance between inputs $\mathbf{x}$ and $\mathbf{x}'$,
%and $l$ and $\sigma_f^2$ are hyperparameters representing the lengthscale and signal variance, respectively.

%The resulting fixed hyperparameters $\theta = \{m_{gp},\,\sigma_n^2,\,\sigma_f^2,\,l\}$
%control the relations between data points within the \ac{GP} model.
%These values can be optimized by training data to match the properties of $f$
%by minimizing the log marginal likelihood function~\citep{Rasmussen2006}.

\subsection{Mapping Under Pose Uncertainty} \label{SS:mapping_under_uncertainty}
To propagate the robot pose uncertainty into our mapping framework,
we \edit{apply} the expected kernel technique of~\citet{Jadidi2019} to planning problems in 3-D setups.
The key idea lies in taking the expectation of \edit{$k$} over \acp{UI}. %the covariance function $k$ over \acp{UI}.
%Let $X \in \mathcal{X}$ be a random variable distributed according to a probability distribution $p(x)$.
\edit{Instead of $X$ being a deterministic location as in  Section \Cref{SS:gaussian_processes}, let $X$ now be a random variable distributed according to a probability distribution $p(x)$.}
The expected covariance function $\tilde{k}$ can be computed as:
\begin{equation}
\begin{aligned}
   \tilde{k} = \mathop{{}\mathbb{E}}[k] =
   \int_{X}kp(x)dx \, \textrm{.} \label{E:modified_kernel}
\end{aligned}
\end{equation}
Assuming a Gaussian distribution for the robot \edit{position} $\mathcal{N}(\mathbf{p}, \, \mathbf{\Sigma})$,
we apply Gauss-Hermite quadrature to efficiently approximate the integral in \Cref{E:modified_kernel} in three dimensions.
Specifically, in this case,
through a change of variable such that $\mathbf{L}\mathbf{L}^\top = \mathbf{\Sigma}$ and $\mathbf{u} = \mathbf{L}^{-1}(\mathbf{x} - \mathbf{p})$, where $\mathbf{L}$ is a lower triangular matrix computed via the Cholesky decomposition:
\begin{equation}
 \tilde{k} = \frac{1}{(2\pi)^\frac{3}{2}} \sum^{\edit{M}}_{i_1 = 1} \sum^{\edit{M}}_{i_2 = 1} \sum^{\edit{M}}_{i_3 = 1} \bar{w}k_{i_{1:3}} \label{E:modified_kernel_3d} \, \textrm{,}
\end{equation}
where
$\edit{M}$ is the number of sample points used in the approximation,
$\bar{w} \triangleq \prod^{3}_{j=1} w_{i_j}$, $u_{i_j}$ corresponds to the roots of the Hermite polynomial $H_n$,
$\mathbf{u}_{i_{1:3}} \triangleq [u_{i_1},u_{i_2},\,u_{i_3}]^\top$,
and $k_{i_{1:3}}$ is the covariance function evaluated at $\mathbf{x}_{i_{1:3}} = \mathbf{L}\mathbf{u}_{i_{1:3}} + \mathbf{p}$.

\section{PLANNING APPROACH} \label{S:planning_approach}
This section overviews our informative planning scheme,
which generates fixed-horizon plans through a combination of a 3-D grid search and evolutionary optimization.
We summarize the key steps of the algorithm,
focusing on our uncertainty-aware utility function
for \ac{GP}-based active mapping scenarios as the main contribution of this paper.
For further details concerning the planning approach itself,
the reader is referred to our previous publications~\citep{Popovic2017ICRA,Popovic2017IROS,Popovic2019}.

\subsection{Trajectories} \label{SS:trajectories}
A polynomial trajectory $\psi$ is represented
by $N$ ordered control waypoints to visit 
$\mathcal{C} = [\mathbf{c}_1, \ldots, \mathbf{c}_N]$ connected using $N-1$ $k$-order spline segments.
Given a reference velocity and acceleration,
we optimize the trajectory for smooth minimum-snap dynamics~\citep{Richter2013},
clamping $\mathbf{c}_1$ as the initial robot position.
In \Cref{E:ipp_problem}, \textsc{measure(\textperiodcentered)}
is defined by computing the spacing of measurement sites
given a constant sensor frequency
and the traveling speed of the robot. %along $\psi$.

\subsection{Algorithm} \label{SS:algorithm}
We plan using a fixed-horizon approach,
alternating between replanning and execution
until the elapsed time $t$ exceeds the budget $B$.
Our replanning strategy (\Cref{A:replan_path}) consists of two steps.
First,
an initial trajectory is obtained through a grid search (Lines~3-7)
based on a coarse set of points $\mathcal{L}$ in the 3-D robot workspace.
In this step, we conduct a sequential greedy search
for $N$ control waypoints in $\mathcal{C}$;
selecting the next-best point $\mathbf{c}^*$ (Line~4)
by evaluating \Cref{E:ipp_problem} over $\mathcal{L}$.
As described in \Cref{SS:utility_definition},
our new utility function (\Cref{E:utility_renyi})
defines the objective $\mathrm{I}($\textperiodcentered$)$ \edit{in \Cref{E:ipp_problem}}
for planning under pose uncertainty.
During this procedure,
for each candidate goal,
the evolution of the robot pose uncertainty $\mathbf{\Sigma}$
\edit{along a trajectory connecting it to the current pose}
is predicted using \ac{SLAM} (Line~5),
as detailed in \Cref{SS:uncertainty_prediction}.
The result is used to update the covariance matrix of the \ac{GP} field model $\mathbf{P}$ (Line~6) 
\edit{via \Cref{E:gp_cov} based on the locations at which new measurements are taken. It is then} added to the \edit{points $\mathcal{C}$ representing the} initial trajectory solution (Line~7).
\begin{algorithm}[!h]
\Crefname{equation}{Eq.}{Equations}
\renewcommand{\algorithmicrequire}{\textbf{Input:}}
\renewcommand{\algorithmicensure}{\textbf{Output:}}
\algrenewcommand\algorithmiccomment[2][\scriptsize]{{#1\hfill\(//\)
\textcolor[rgb]{0.4, 0.4, 0.4}{#2} }}
\begin{algorithmic}[1]
  \Require Current covariance matrix of the \ac{GP} field model $\mathbf{P}$, number of control waypoints $N$, grid points $\mathcal{L}$, initial position $\mathbf{c}_1$, robot pose $(\mathbf{p},\,\mathbf{\Sigma})$
  \Ensure Waypoints defining next polynomial plan $\mathcal{C}$
  \State $\mathbf{P}' \gets \mathbf{P}$; $(\mathbf{p}',\,\mathbf{\Sigma'}) \gets (\mathbf{p},\,\mathbf{\Sigma})$ \Comment{Create local copies.}
  \State $\mathcal{C} \gets \mathbf{c}_1$ \Comment{Initialize control points.}
  \While {$N \geq |\mathcal{C}|$}
   \State $\mathbf{c}^* \gets$ Select viewpoint in $\mathcal{L}$ using 
   \Cref{E:ipp_problem}
   \State ($\mathbf{p}'$, $\mathbf{\Sigma}'$) $\gets$ \Call{predict\_motion}{$\mathbf{p}'$, $\mathbf{\Sigma}'$, $\mathbf{c}^*$}
  %\State $Z' \gets$ \Call{update\_map}{$Z'$, $\textbf{c}^*$}
  %\Comment{Predict a measurement from this point.}
   \State $\mathbf{P}' \gets$ \Call{predict\_measurement}{$\mathbf{P}'$, $\mathbf{p}'$, $\mathbf{\Sigma}'$}
  %  \Comment{From this point.}
   \State $\mathcal{C} \gets \mathcal{C} \cup \mathbf{c}^*$
  \EndWhile
  \State $\mathcal{C} \gets$ \Call{cmaes}{$\mathcal{C}$, $\mathbf{P}$, $\mathbf{p}$, $\mathbf{\Sigma}$}
\Comment{Optimize control points \edit{using \Cref{E:ipp_problem}}.}
\end{algorithmic}
\caption{\textsc{replan\_path} procedure}\label{A:replan_path}
\end{algorithm}

The second replanning step (Line~8)
refines the coarse grid search output
\edit{for $\mathcal{C}$ using \Cref{E:ipp_problem}.
This is done by computing $\mathrm{I}($\textperiodcentered$)$
for a sequence of measurements taken along the corresponding trajectory.
For prediction,
we apply the same principles as described above in Lines~5 and 6.
An important benefit of our two-step approach
is that the informed initialization procedure exploiting the grid-based solution
effectively speeds up the convergence of the optimizer,
making it suitable for quickly finding solutions on computationally limited systems.

Note that the optimization step is agnostic to the actual method considered;
specifically, in this work, we employ the Covariance Matrix Adaptation Evolution Strategy (CMA-ES)~\citep{Hansen2006}.
The CMA-ES is a generic global derivative-free optimizer based on evolutionary algorithms
that has been successfully applied to solve high-dimensional, nonlinear, non-convex
problems in a continuous domain~\citep{Popovic2017ICRA,Popovic2017IROS,Hitz2017,Hansen2006}.
Essentially,
it operates by iteratively evolving candidate solutions drawn from a search distribution whose parameters are adapted over time.
Our choice of this routine is motivated by the nonlinearity of the objective space in \Cref{E:ipp_problem} as well as by the above mentioned results.
For further details, the reader is referred to the in-depth review of~\citet{Hansen2006}.}

\subsection{Utility Definition} \label{SS:utility_definition}
\edit{
We introduce a new utility, or information gain, function $\mathrm{I}($\textperiodcentered$)$ in \Cref{E:ipp_problem} which directly relates the uncertainty of both the robot pose and the field map without the need for additional hand-tuned scaling parameters. Our utility function is derived from the R\'{e}nyi entropy, which for a multivariate Gaussian distribution is given by \citep{Golshani2010}:
\begin{equation}
  \mathrm{H}_{\alpha}\!\left(\mathbf{X}\right) =
    \frac{1}{2}\log{\left(\left\lvert\left(2\pi\alpha^{\frac{1}{\alpha-1}}\right)\mathbf{P}\right\rvert\right)} \, \textrm{,}
 \label{E:gaussian_renyi}
\end{equation}
where $\alpha \in [0,1) \cup (1, \infty)$ is a free parameter. Note that as $\alpha \rightarrow 1$, the R\'{e}nyi entropy converges to the Shannon entropy in the limiting case.

To improve computational tractability, we approximate the determinant in \Cref{E:gaussian_renyi} using the trace of the covariance matrix~\citep{Costante2017,Hitz2017,Popovic2017IROS}. Dropping the constant terms then gives:
\begin{equation}
   \mathrm{H}_{\alpha}\!\left(\mathbf{X}\right)\appropto\hat{\mathrm{H}}_{\alpha}\!\left(\mathbf{X}\right) =
    \log{\left(\Tr{\left(\mathbf{P}\right)} \, \alpha^{\frac{1}{\alpha-1}}\right)} \, \textrm{,}
  \label{E:gp_renyi}
\end{equation}
where $\Tr($\textperiodcentered$)$ denotes the matrix trace, and the covariance matrix $\mathbf{P}$ is obtained using the expected kernel in \Cref{E:modified_kernel} for mapping under uncertainty.

Our key insight is to relate the free parameter $\alpha$ to the predicted pose uncertainty $\mathbf{\Sigma}$ along a candidate trajectory such that the expected information gain reduces when localization uncertainty is high. Noting that $\mathrm{H}_{\alpha}$ is strictly non-increasing in $\alpha$, this intuition can be captured by setting $\alpha$ to:
\begin{equation}
    \alpha\!\left(\mathbf{\Sigma}\right)=1+\frac{1}{\mathrm{Tr}\left(\mathbf{\Sigma}\right)},
    \label{E:alpha_func}
\end{equation}
and computing information gain as:
\begin{equation}
  \mathrm{I}_{\alpha(\mathbf{\Sigma})}(\mathbf{p}) = \hat{\mathrm{H}}(\mathbf{P}^-) - \hat{\mathrm{H}}_{\alpha(\mathbf{\Sigma})}(\mathbf{P}^+) \, \textrm{,}
 \label{E:utility_renyi}
\end{equation}
where the superscripts $-$ and $+$ denote the prior and posterior covariance matrix $\mathbf{P}$ of the \ac{GP} model, respectively. Similar to our approximation in \Cref{E:gp_renyi}, the first term on the right hand side in \Cref{E:utility_renyi} is computed using the trace of the \ac{GP} covariance. This approximates the Shannon entropy of the model before executing the candidate trajectory. The second term approximates the R\'{e}nyi entropy of the model given the predicted observations, which is shaped according to the associated pose uncertainties by \Cref{E:alpha_func}. Thus, \Cref{E:utility_renyi} bears similarities to the mutual information, with the key difference that we are able to explicitly discount the information gain according to the predicted localization uncertainty along the candidate path.

It is worth noting that other functions may be used to relate $\alpha$ to the localization uncertainty. For example, \citet{Carrillo2018} also suggest using the determinant or the maximum eigenvalue of $\mathbf{\Sigma}$ in place of the trace in \Cref{E:alpha_func}. Nevertheless, modulating the R\'{e}nyi entropy provides a formal way in which to incorporate the localization uncertainty while maintaining a utility function that is primarily computed over the \ac{GP} field map uncertainty.
}

\subsection{Uncertainty Prediction} \label{SS:uncertainty_prediction}
A key requirement for predictive planning is propagating
the robot localization uncertainty for a candidate action (Line~5 of \Cref{A:replan_path}).
\edit{We consider this problem for active sensing
in initially unknown and known environments.}

\subsubsection{\edit{Unknown Environments}} \label{SSS:unknown_environments}

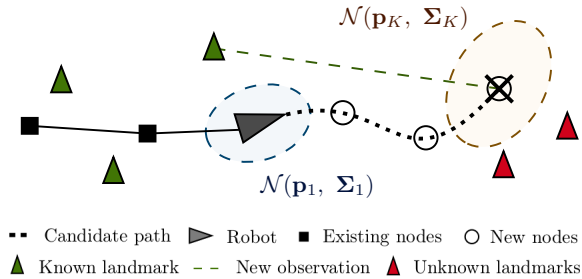
\begin{figure}[!h]
\centering
  \resizebox{0.90\columnwidth}{!}{
  \tikzset{every picture/.style={line width=0.75pt}} %set default line width to 0.75pt        

\begin{tikzpicture}[x=0.75pt,y=0.75pt,yscale=-1,xscale=1]

%Shape: Ellipse [id:dp1619038137379284] 
\draw  [color={rgb, 255:red, 0; green, 68; blue, 110 }  ,draw opacity=1 ][fill={rgb, 255:red, 9; green, 88; blue, 160 }  ,fill opacity=0.05 ][dash pattern={on 4.5pt off 4.5pt}] (180.69,124.68) .. controls (177.54,114.57) and (186.85,102.67) .. (201.47,98.12) .. controls (216.1,93.56) and (230.52,98.07) .. (233.67,108.18) .. controls (236.82,118.3) and (227.51,130.2) .. (212.88,134.75) .. controls (198.26,139.31) and (183.84,134.8) .. (180.69,124.68) -- cycle ;
%Straight Lines [id:da26553437980014716] 
\draw [color={rgb, 255:red, 53; green, 94; blue, 5 }  ,draw opacity=1 ] [dash pattern={on 4.5pt off 4.5pt}]  (184.25,77.73) -- (331.97,98.57) ;

%Shape: Ellipse [id:dp9051066970200505] 
\draw  [color={rgb, 255:red, 127; green, 80; blue, 3 }  ,draw opacity=1 ][fill={rgb, 255:red, 245; green, 166; blue, 35 }  ,fill opacity=0.05 ][dash pattern={on 4.5pt off 4.5pt}] (314.48,126.49) .. controls (302.89,119.89) and (301.23,100.96) .. (310.76,84.19) .. controls (320.3,67.43) and (337.43,59.19) .. (349.02,65.78) .. controls (360.61,72.38) and (362.28,91.32) .. (352.74,108.08) .. controls (343.2,124.84) and (326.07,133.08) .. (314.48,126.49) -- cycle ;
%Shape: Triangle [id:dp7879638468473792] 
\draw  [fill={rgb, 255:red, 65; green, 117; blue, 5 }  ,fill opacity=1 ] (105.25,87) -- (110.5,100.47) -- (100,100.47) -- cycle ;
%Shape: Triangle [id:dp12638487757675154] 
\draw  [fill={rgb, 255:red, 65; green, 117; blue, 5 }  ,fill opacity=1 ] (132.25,134) -- (137.5,147.47) -- (127,147.47) -- cycle ;
%Shape: Triangle [id:dp10973414820655158] 
\draw  [fill={rgb, 255:red, 65; green, 117; blue, 5 }  ,fill opacity=1 ] (184.25,70) -- (189.5,83.47) -- (179,83.47) -- cycle ;
%Curve Lines [id:da08491519866901809] 
\draw [line width=1.5]  [dash pattern={on 1.69pt off 2.76pt}]  (221.7,111.99) .. controls (278.08,98.73) and (279.97,155.57) .. (331.97,98.57) ;

%Shape: Triangle [id:dp05043822385629826] 
\draw  [fill={rgb, 255:red, 74; green, 74; blue, 74 }  ,fill opacity=1 ] (220.7,111.99) -- (199.47,126.47) -- (195,111.82) -- cycle ;
%Shape: Circle [id:dp5292623587228731] 
\draw   (245,111.04) .. controls (245,107.7) and (247.7,105) .. (251.04,105) .. controls (254.38,105) and (257.08,107.7) .. (257.08,111.04) .. controls (257.08,114.38) and (254.38,117.08) .. (251.04,117.08) .. controls (247.7,117.08) and (245,114.38) .. (245,111.04) -- cycle ;
%Shape: Circle [id:dp03965748273249947] 
\draw   (288,124.04) .. controls (288,120.7) and (290.7,118) .. (294.04,118) .. controls (297.38,118) and (300.08,120.7) .. (300.08,124.04) .. controls (300.08,127.38) and (297.38,130.08) .. (294.04,130.08) .. controls (290.7,130.08) and (288,127.38) .. (288,124.04) -- cycle ;
%Shape: Circle [id:dp45273293547088267] 
\draw   (325.93,98.57) .. controls (325.93,95.23) and (328.63,92.53) .. (331.97,92.53) .. controls (335.3,92.53) and (338.01,95.23) .. (338.01,98.57) .. controls (338.01,101.9) and (335.3,104.61) .. (331.97,104.61) .. controls (328.63,104.61) and (325.93,101.9) .. (325.93,98.57) -- cycle ;
%Shape: Triangle [id:dp2731597018382481] 
\draw  [fill={rgb, 255:red, 208; green, 2; blue, 27 }  ,fill opacity=1 ] (367.25,111) -- (372.5,124.47) -- (362,124.47) -- cycle ;
%Shape: Triangle [id:dp8836439439844626] 
\draw  [fill={rgb, 255:red, 208; green, 2; blue, 27 }  ,fill opacity=1 ] (334.25,131) -- (339.5,144.47) -- (329,144.47) -- cycle ;
%Shape: Square [id:dp8432562069681819] 
\draw  [color={rgb, 255:red, 0; green, 0; blue, 0 }  ,draw opacity=1 ][fill={rgb, 255:red, 0; green, 0; blue, 0 }  ,fill opacity=1 ] (84.79,113.71) -- (93.04,113.71) -- (93.04,121.96) -- (84.79,121.96) -- cycle ;
%Shape: Square [id:dp2932737302114914] 
\draw  [fill={rgb, 255:red, 0; green, 0; blue, 0 }  ,fill opacity=1 ] (145.79,117.71) -- (154.04,117.71) -- (154.04,125.96) -- (145.79,125.96) -- cycle ;
%Straight Lines [id:da7067985624944856] 
\draw    (88.92,117.83) -- (149.92,121.83) ;

%Straight Lines [id:da16050420158302736] 
\draw    (149.92,121.83) -- (197.25,119.9) ;

\draw  [line width=1.5]  (325.01,91.75) -- (338.94,105.59)(338.89,91.7) -- (325.06,105.63) ;

% Text Node
\draw (120,142) node  [align=left] {};
% Text Node
\draw (238,149) node [scale=0.9] [align=left] {$\displaystyle \textcolor[rgb]{0,0.07,0.24}{\mathcal{N}(\mathbf{p}_{1} ,\ \mathbf{\Sigma }_{1})}$};
% Text Node
\draw (283,62) node [scale=0.9] [align=left] {$\displaystyle \textcolor[rgb]{0.29,0.09,0}{\mathcal{N}(\mathbf{p}_{K} ,\ \mathbf{\Sigma }_{K})}$};

\end{tikzpicture}} \\
  \vspace*{1.5mm}
  \includegraphics[width=0.9\columnwidth]{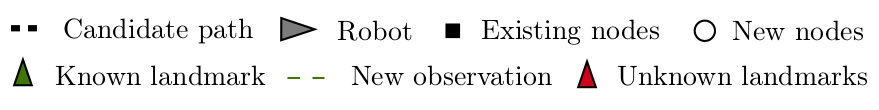}
  \caption{Uncertainty prediction using graph \ac{SLAM} with point landmarks.
  The dashed line is a candidate path to `X'.
  Example uncertainty ellipses at initial and final positions are shown.
  To predict the robot pose evolution,
  new nodes are interpolated along the path to extend the current graph.
  %The green dotted line depicts the re-observation of a known landmark,
  %while new landmarks in unknown space are ignored.
  The extended graph is optimized
  to estimate the final pose $\mathcal{N}(\mathbf{p}_K,\,\mathbf{\Sigma}_K)$
  for planning.}\label{F:uncertainty_prediction}
\end{figure}

In unknown environments,
we consider a solution to this task assuming graph \ac{SLAM}
using odometry and point landmark observations as constraints~\citep{Sola2017}.
\Cref{F:uncertainty_prediction} schematizes a 2-D example.
To predict the localization uncertainty along a possible path (dashed line),
the graph is simply extended from the current pose $\mathcal{N}(\mathbf{p}_1,\,\mathbf{\Sigma}_1)$.
The trajectory is interpolated at a fixed frequency
to add $K-1$ odometry constraints (hollow circles) in the extended graph,
giving rise to the sequence
$\{\mathcal{N}(\mathbf{p}_1,\,\mathbf{\Sigma}_1), \ldots, \mathcal{N}(\mathbf{p}_K,\,\mathbf{\Sigma}_K)\}$.
For each consecutive node pair,
we apply control noise drawn from a Gaussian distribution
whose variance is proportional to the control input magnitude.
This reflects the fact that longer motion steps
are likely to be associated with higher actuation errors. %associated with them.

To address potential loop closures,
we simulate re-observations to the known landmarks (green triangles) maintained in the current graph \edit{given the predicted robot pose and sensor \ac{FoV}}.
In unknown space,
we assume that no new landmarks will be detected,
\edit{and thus, the robot pose uncertainty will grow as a result of the control noise}.
The resulting graph is then solved using QR factorization
to obtain $\mathbf{\Sigma}_K$.
Future work will consider extending these ideas to other \ac{SLAM} systems.
%and sparsification methods~\citep{Vallve2018}
%for real-time application in larger areas.

\subsubsection{\edit{Known Environments}} \label{SSS:known_environments}

Similarly, in a known environment,
the uncertainty can be predicted
assuming a \ac{MCL} approach.
\edit{This method uses
a motion model and the expected sensor measurements obtained
to estimate the robot state by propagating the distribution of particles in a particle filter.}
In \Cref{SS:experiments},
this method is applied
to localize a robot with a laser scanner in an indoor area.

Putting together our ideas in mapping and planning in a single framework,
the \ac{GP} field model with \acp{UI} accounts for the pose uncertainty in the received observations,
while the proposed utility function accounts for the pose uncertainty in potential future measurements.
This coupling allows us to achieve robust performance
in \ac{GP}-based mapping scenarios.

\section{EXPERIMENTAL RESULTS} \label{S:experimental_results}
In this section,
our approach is evaluated in simulation
by comparing it to different strategies for planning and mapping.
We then show proof of concept
by using it to map temperature
using a ground robot in an indoor environment.

\subsection{Comparison of Planning Methods} \label{SS:evaluation_planning}
First,
the aim is to evaluate our new utility function
by comparing it against existing strategies.
To focus on examining \edit{our utility function and planning performance in particular},
all methods in this sub-section use the \ac{GP}-based approach
with \acp{UI} for mapping under pose uncertainty (\Cref{SS:mapping_under_uncertainty}).
\edit{The experiments are executed in MATLAB running on an Intel $1.8$\,GHz computer with $16$\,GB of RAM.}

Our experimental setup considers \edit{5 different} $5$\,m$\times$\,$5$\,m$\times$\,$4$\,m Gaussian Random Field \edit{environments for mapping,
assumed to be initially unknown}.
We use a $0.25$\,m$\times$\,$0.25$\,m$\times$\,$1$\,m resolution grid
and apply the isotropic \ac{SE} kernel
with hyperparameters %$\theta = \{17.61, \,2.13 \times 10^{-11}, \,1141.38, \,2.18\}$
trained by minimizing log marginal likelihood \edit{in each environment}~\citep{Rasmussen2006}.
\edit{For mapping with \acp{UI}, the modified kernel} in \Cref{E:modified_kernel}
is estimated using Gauss-Hermite quadrature with 5 points.
During a mission,
field measurements are taken at $0.25$\,Hz using a point-based sensor
centered on
\edit{a 3 degrees of freedom robot
moving as a point mass along the $(x,y,z)$ axes}.
\edit{These new measurements} are added to our \ac{GP} model
as observed input training points
to achieve uncertainty reduction
as exploration takes place.

\begin{figure}[!h]
\centering
 \begin{subfigure}[]{0.24\textwidth}
  \includegraphics[width=0.76\columnwidth]{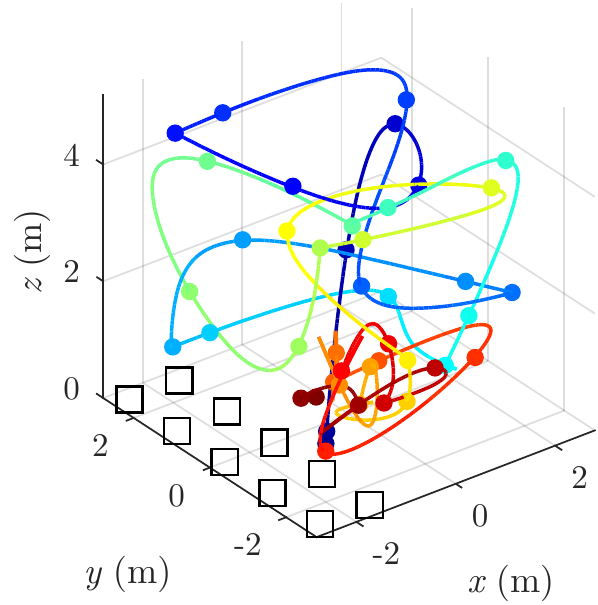}
  \hspace{1mm}
  \raisebox{2mm}{\includegraphics[width=0.17\columnwidth]{images/teaser_traj_colorbar.pdf}}
  \caption{} \label{SF:results_planning_example_traj}
 \end{subfigure}\hfill%
 \begin{subfigure}[]{0.24\textwidth}
  \includegraphics[width=0.76\columnwidth]{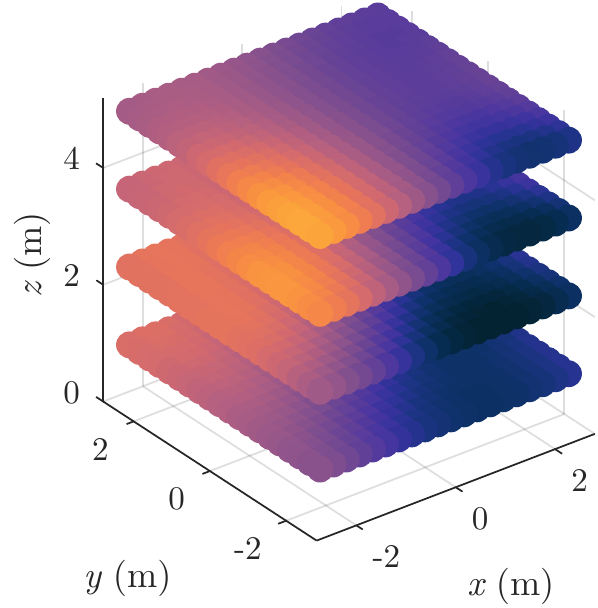}
  \hspace{1mm}
  \raisebox{2mm}{\includegraphics[width=0.17\columnwidth]{images/teaser_map_colorbar.pdf}}
  \caption{} \label{SF:results_planning_example_map}
 \end{subfigure}\hfill%
   \caption{Example result using \edit{our CMA-ES-based replanning framework and our} R\'enyi-based utility function.
   (a) shows the trajectory traveled by the robot.
   The squares indicate point landmarks \edit{on the ground} used for localization.
   The spheres represent sites where measurements are taken
   to produce the final field map in (b).
   By balancing between gathering new information and keeping the landmarks in view,
   our planner achieves 1.86 lower total field map RMSE compared to the spiral path shown in \Cref{F:teaser}.
   % Renyi: 1.11. Spiral: 2.06.
   }\label{F:results_planning_example}
\end{figure}

To emulate an aerial robot setup,
10 \edit{visual 3-D} point landmarks are placed on the ground
$1$\,m below the target field
\edit{and are distributed over one side of the space, as shown in \Cref{SF:teaser_trajectory,SF:results_planning_example_traj}}.
For \ac{SLAM}, the robot is equipped with a downward-facing camera
with $(47.9\degree,\,36.9\degree)$ \ac{FoV}.
\edit{Landmark detection is done based on a pinhole projection camera model with standard deviations of $1.0$\,px and $0.1$\,m for measurement errors in pixel and depth.}
Our framework uses the graph \ac{SLAM} algorithm~\citep{Sola2017} \edit{implemented in the \ac{SLAM} Toolbox\footnote{\url{github.com/joansola/slamtb}},}
with the approach described in \edit{\Cref{SSS:unknown_environments}} for \edit{predicting the robot localization uncertainty in unknown environments}.
We sample trajectories at $0.5$\,Hz to simulate control actions \edit{using an odometry motion model},
applying a coefficient of $0.01$ in all three co-ordinate dimensions
to scale the control noise variance.
For planning,
covariance matrices from the graph are extracted
through the computations developed by~\citet{Kaess2009}.

Our \edit{two-step CMA-ES-based replanning framework (\Cref{SS:algorithm}) with the} R\'enyi-based utility function (\Cref{SS:utility_definition})
is compared against itself using different objectives:
(a) field map uncertainty reduction only (\Cref{E:utility_renyi} using standard Shannon's entropy \edit{in the map posterior});
(b) field map uncertainty reduction over time (rate), as in our previous works~\citep{Popovic2017ICRA,Popovic2017IROS,Popovic2019};
and (c) a linear composite of the map and pose uncertanties weighting both objectives equally based on their upper bounds, \edit{tuned according to the approach of~\citet{Bourgault2002}}.
As planning benchmarks,
we also study:
the rapidly exploring information gathering tree (RIG-tree) \edit{using objectives of (d) uncertainty reduction only and (e) our R\'enyi-based utility,
as well as (f)} random waypoint selection. 
\edit{The former is a state-of-the-art sampling-based planning algorithm
introduced by~\citet{Hollinger2014}
that incrementally extends a geometric tree
to find the best information gathering trajectory.
We consider this benchmark to assess our two-step replanning approach
and, in (f), show the applicability of our generic concepts in uncertainty-aware active sensing
with different planners.}
A $150$\,s budget $B$ is specified for all methods.
To evaluate mapping performance,
we quantify uncertainty with the \ac{GP} covariance trace $\Tr(\mathbf{P})$
and accuracy with \ac{RMSE} with respect to the ground truth map.
Similarly, for planning,
our measures are the robot covariance trace $\Tr(\mathbf{\Sigma})$
and the pose \ac{RMSE} with respect to the ground truth trajectory.
Intuitively,
lower values using all metrics signify better performance.
\begin{figure*}[!h]
\begin{subfigure}[]{0.485\textwidth}
\centering
  \includegraphics[width=0.76\columnwidth]{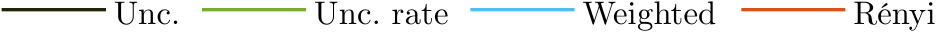} \\
  \vspace*{0.1mm}
  \includegraphics[width=\columnwidth]{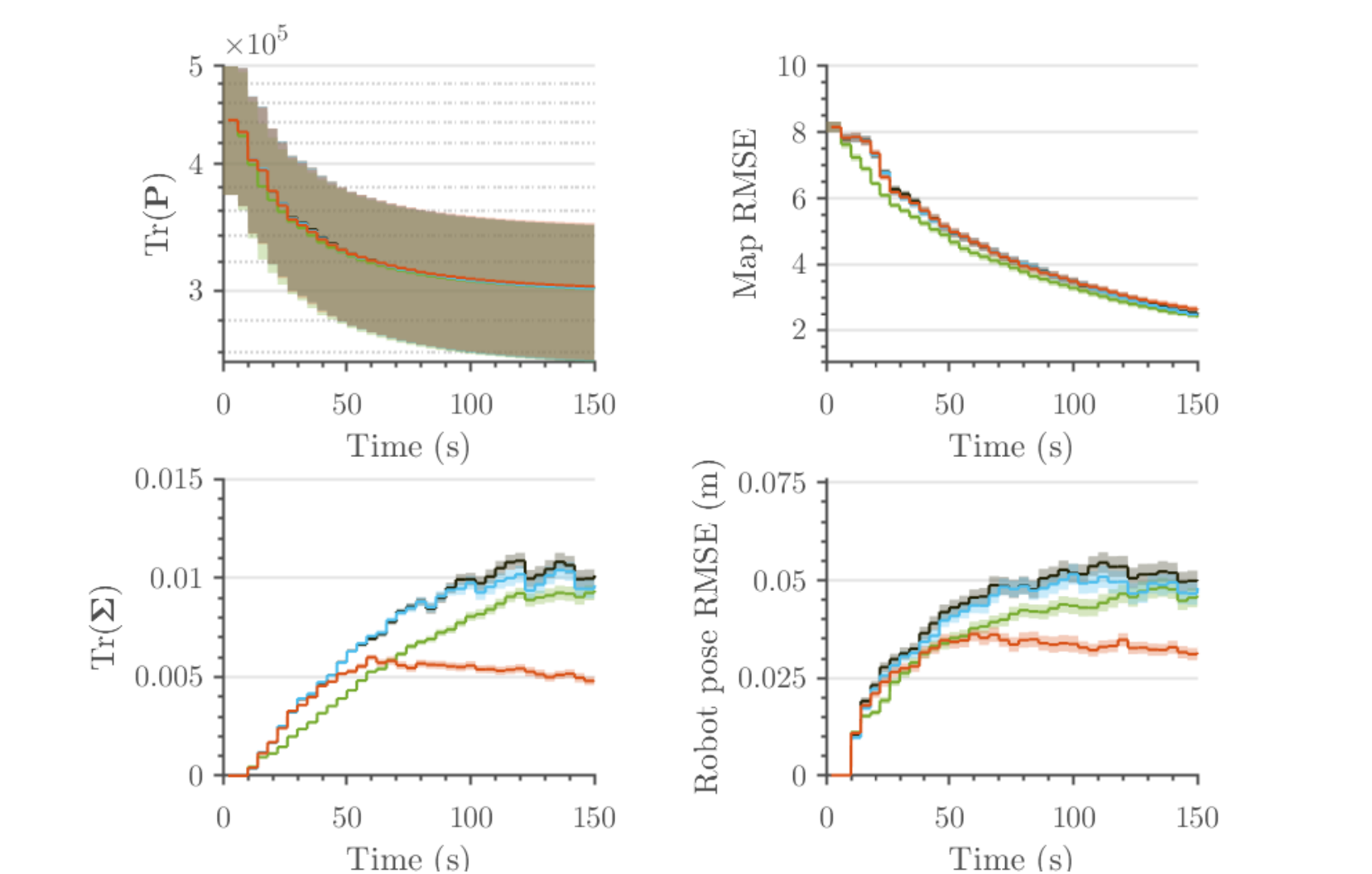} \caption{} \label{SF:results_planning_objectives}
\end{subfigure}
\begin{subfigure}[]{0.485\textwidth}
\centering
  \includegraphics[width=0.9\columnwidth]{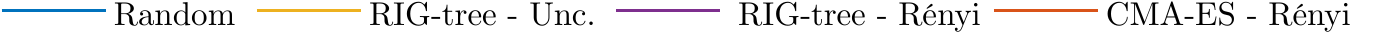} \\
  \vspace*{0.1mm}
  \includegraphics[width=\columnwidth]{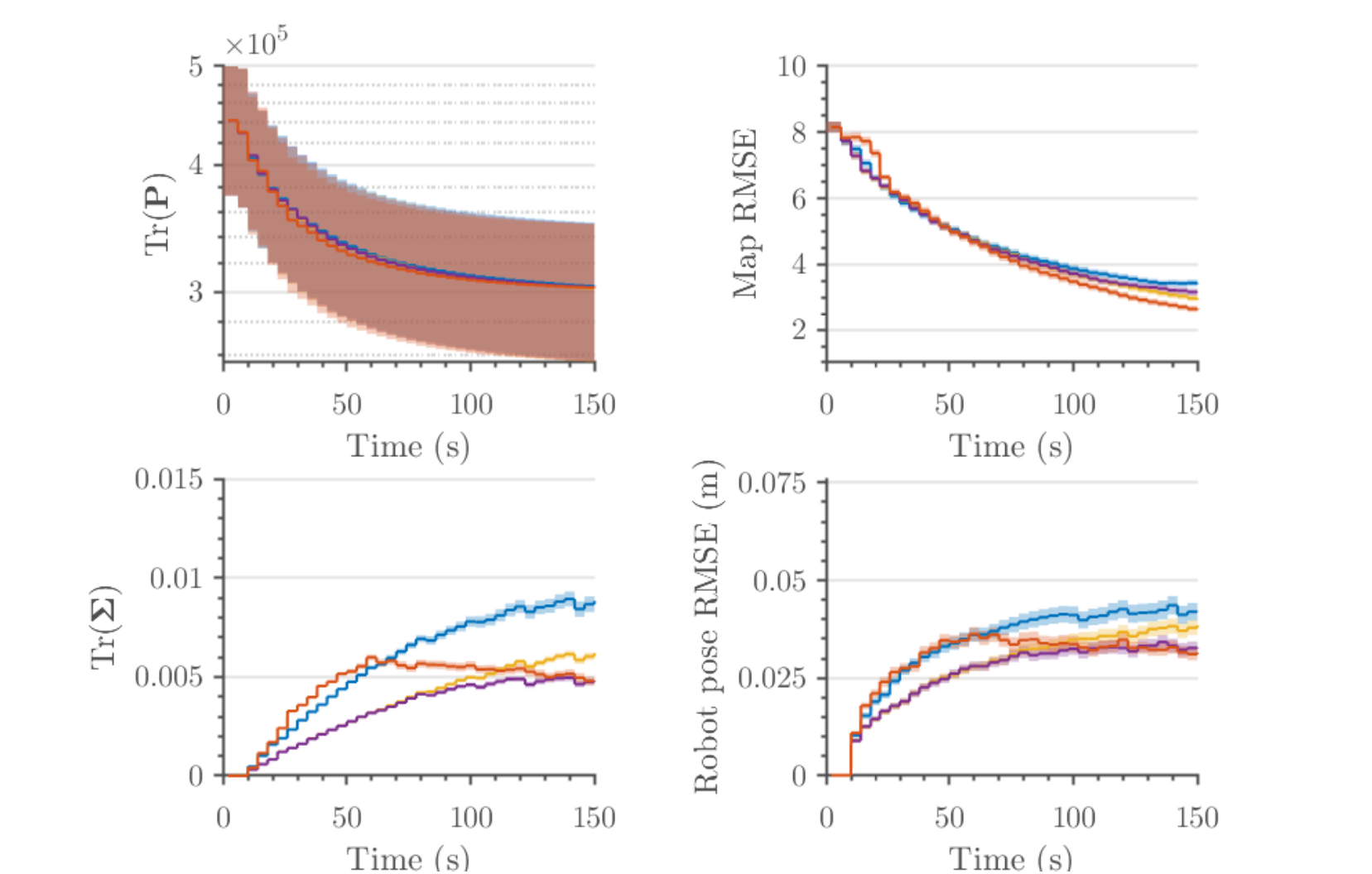} \caption{} \label{SF:results_planning_benchmarks}
\end{subfigure}
   \caption{Comparison of our two-step CMA-ES-based replanning framework using our R\'enyi-based utility function
   against \edit{(a) our framework using different information objectives and (b) different planning benchmarks.
   All methods are given a fixed budget of $150$\,s and use} \ac{GP}-based \edit{field} mapping with \acp{UI}.
   The solid lines represent means over $250$ trials. The shaded regions show 95\% confidence bounds.
   By considering both the robot pose and field map uncertainties,
   our \edit{utility function} more quickly achieves higher-quality mapping (top \edit{row})
   with improved localization (bottom \edit{row}).
   Note that $\Tr(\mathbf{P})$ is on a logarithmic scale\edit{, and that its variance is large due to differences between the field mapping environments considered in the trials.}
   }\label{F:results_planning}
\end{figure*}

\edit{In all environments,} the starting robot position is ($2$\,m, $2$\,m, $1$\,m)
with no prior \edit{localization nor field map information}.
For trajectory optimization,
the reference velocity and acceleration are $1.5$\,m$/$s and $3$\,m$/$s$^2$
using polynomials of order $k = 12$.
In our planner,
we define polynomials with $N = 4$ waypoints
and use a uniformly spaced 27-point lattice for the 3-D grid search.
In RIG-tree, we associate control waypoints with vertices,
and form polynomials by tracing the parents of leaf vertices to the root.
The finite-horizon replanning procedure follows the approach of~\citet{Popovic2019}
with the uncertainty-only objective
and a branch expansion step-size of $5$\,m empirically set for best performance based on \edit{a search over a discrete range of values.
We set $60$ tree-sampling iterations
to obtain the same $\sim 76$\,s average replanning time
as required by our replanning routine for a single trajectory.}
In the random planner,
random destinations are sampled in the \edit{workspace}
and a trajectory is generated by connecting them sequentially to the current robot position.
We consider $4$ waypoints per plan
to ensure that the trajectory lengths are fairly comparable to our method.

\edit{We conduct $50$ simulations in each environment, giving a total of $250$ trials.}
\Cref{F:results_planning} shows how the metrics evolve for each \edit{planning method}.
As expected, \edit{in \Cref{SF:results_planning_benchmarks}},
informed strategies perform better than the random benchmark (blue)
as they are guided by planning objectives.
In \Cref{SF:results_planning_objectives},
the uncertainty-based (black, green) and weighted (cyan) utility functions
yield map uncertainty and error reduction rates
similar to our R\'enyi-based objective (red).
However,
our method significantly improves upon the robot pose estimation in the same scenario;
confirming that it effectively
trades off between gathering information and maintaining good localization
given the known landmarks.
This cannot be done using manually tuned parameters (cyan)
as the variability of the pose uncertainty is much lower
compared to that of the map.
In terms of localization,
the uncertainty-only function (black) performs worst
as it does not exploit any knowledge of the trajectory dynamics.

\edit{Similarly, in \Cref{SF:results_planning_objectives},
the RIG-tree variant using our R\'enyi-based (purple) objective provides better localization when compared against planning for map uncertainty reduction only (yellow).
This further validates the expected behaviour of our utility function, as described above, using a different replanning routine.
Interestingly, compared to our replanning approach (red),}
the RIG-tree scores comparatively well on the localization metrics
whereas the rate at which the map quality improves is limited.
This likely relates to the step-size parameter required by the algorithm,
which sets the range of navigation achievable during the mission.
By traveling trajectories with shorter steps,
the robot re-observes landmarks more frequently
at the cost of restricted exploration;
\edit{leading to a poorer final field reconstruction}.

An example result using \edit{our CMA-ES-based framework and our} proposed utility function
is shown in \Cref{F:results_planning_example}.
\Cref{SF:results_planning_example_traj} confirms that
the robot successfully explores the environment
while re-visiting known landmarks to stay well-localized.
With a total \ac{RMSE} of 1.11,
the map output in this instance (\Cref{SF:results_planning_example_map})
is 1.86 times more accurate than the one produced by the na\"{i}ve spiral path in \Cref{F:teaser},
thus justifying our informative planning strategy.

\subsection{Evaluation of Field Mapping Under Uncertainty} \label{SS:evaluation_mapping}
Next, the aim is to assess the benefits of mapping under the robot pose uncertainty
by evaluating the effects of incorporating \acp{UI} in the \ac{GP} field model.
We consider the same simulation setup as described above
using our \edit{CMA-ES-based replanning framework and} proposed R\'enyi-based utility function \edit{as the information objective}.
We conduct $50$ trials \edit{in each of the 5 environments}
(a) with and (b) without applying the modified kernel presented in \Cref{SS:mapping_under_uncertainty}.
Note that (b) corresponds to a standard \ac{GP} mapping approach \edit{without \acp{UI}} as a benchmark.
\begin{figure}[!h]
\centering
  \includegraphics[width=0.45\columnwidth]{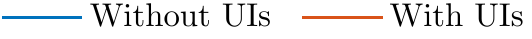} \\
  \vspace*{1mm}
  \includegraphics[width=\columnwidth]{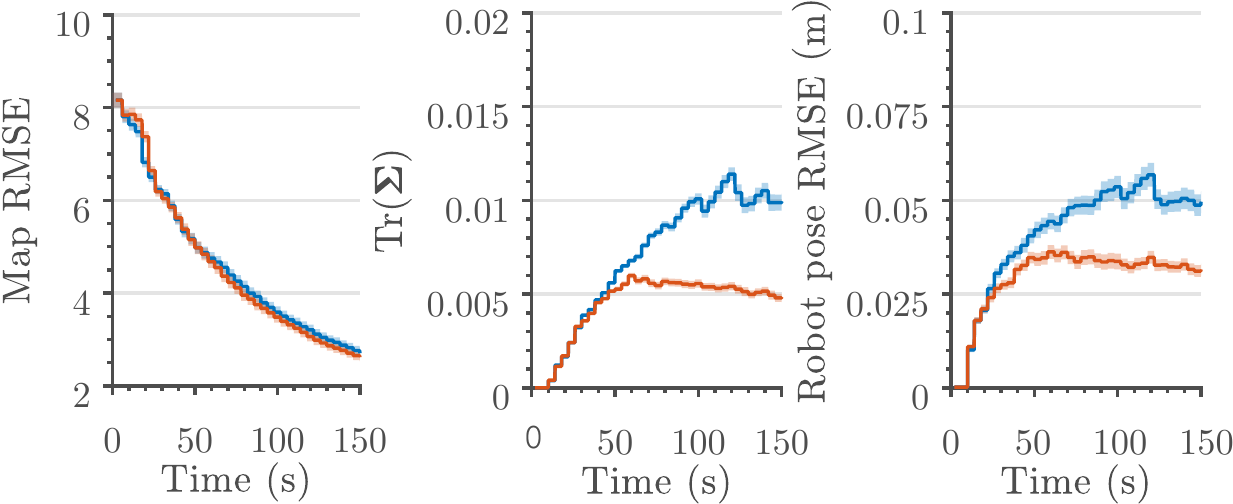}
   \caption{Comparison of field mapping with and without uncertain inputs (UIs)
   using our \edit{CMA-ES-based replanning framework and} R\'enyi-based utility function for a fixed budget of $150$\,s.
   The solid lines represent means over \edit{$250$} trials. The shaded regions show 95\% confidence bounds.
   By accounting for the robot pose uncertainty,
   our approach achieves more conservative mapping behavior (middle, right) with higher accuracy (left).
   }\label{F:results_mapping}
\end{figure}

Our results are depicted in \Cref{F:results_mapping}.
Note that we omit the map uncertainty metric
as the variance scales using the two approaches are not comparable.
The plots confirm that our approach with \acp{UI} (red)
presents more conservative exploratory behavior than the benchmark (blue)
while also yielding more accurate field reconstructions.
This is because our modified kernel can handle localization errors
to build maps with better consistency and quality for reliable planning.
\Cref{F:teaser} shows a visualization that supports this result.

\edit{In summary,
\Cref{F:teaser,F:results_mapping} demonstrate that accounting for \acp{UI} in the field model improves mapping performance, whereas \Cref{F:results_planning} shows that incorporating the robot pose uncertainty in our utility function yields much better localization.
By propagating the pose uncertainty into both the mapping and planning tasks,
our proposed active sensing framework offers brings these benefits together to achieve robust mapping behaviour.}

\subsection{Proof of Concept} \label{SS:experiments}
\begin{figure}[!h]
\centering
  \begin{subfigure}{.497\columnwidth}
    \includegraphics[width=\columnwidth]{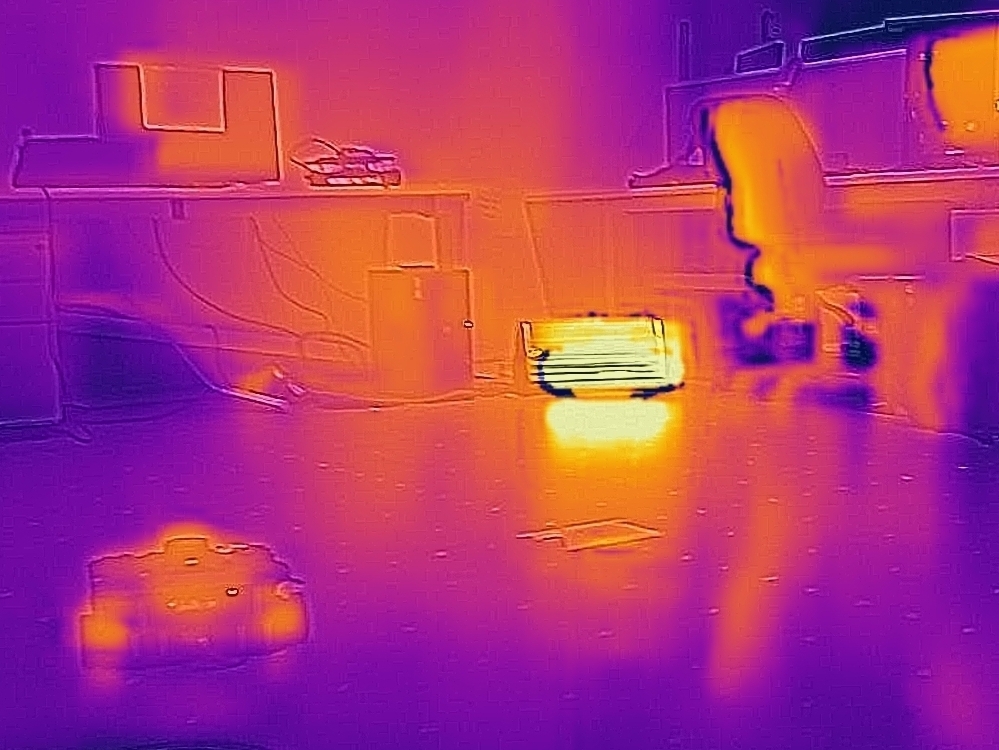}%
    \caption{}%
    \label{SF:setup_side}%
  \end{subfigure}\hfill%
  \begin{subfigure}{.497\columnwidth}
    \includegraphics[width=\columnwidth]{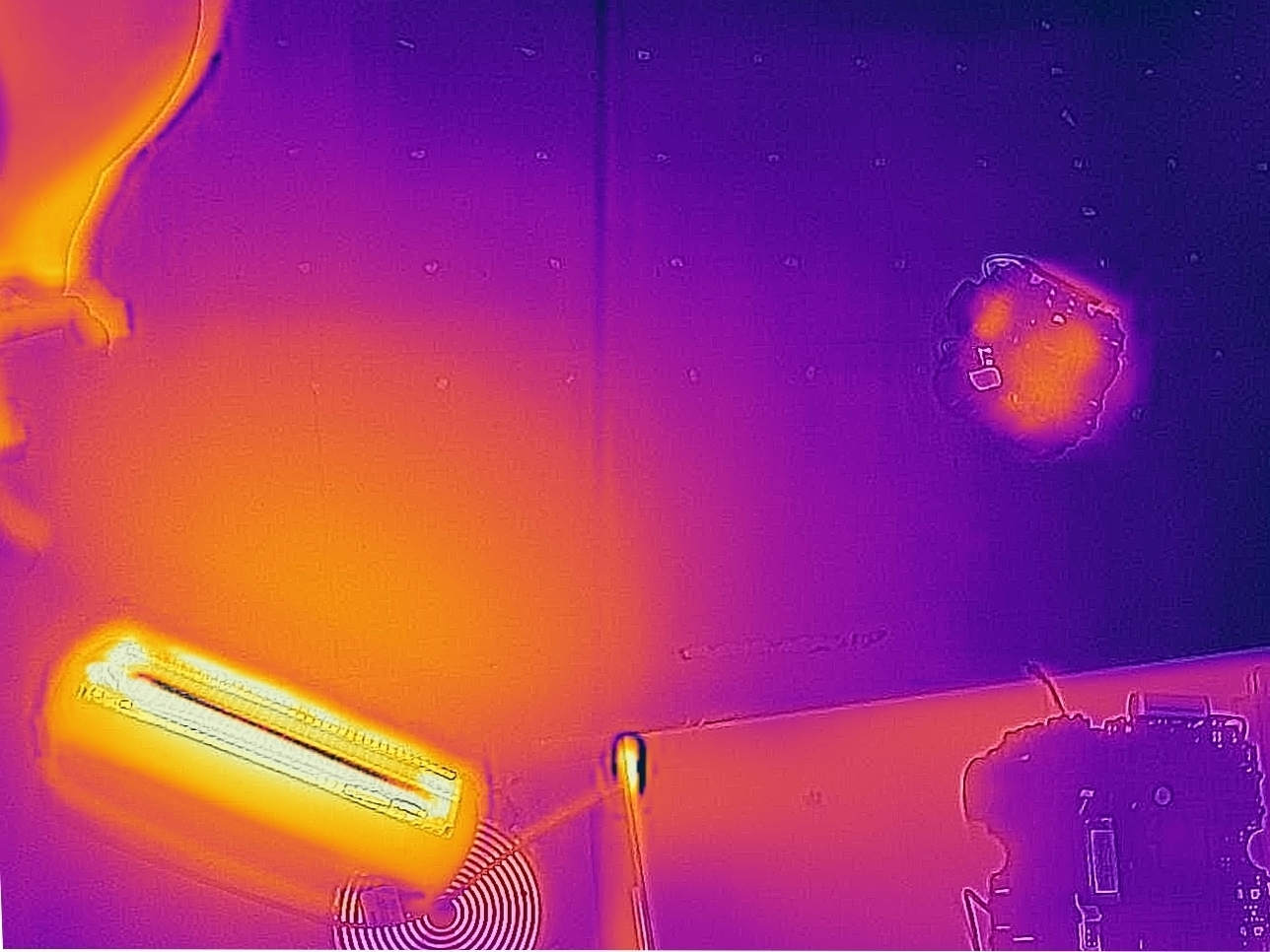}%
    \caption{}%
  \label{SF:setup_environment}%
  \end{subfigure}%
   \caption{Thermal imagery of our experimental setup
   from side (a) and aerial (b) viewpoints.
   The robot and radiator are visible.
   Yellower shades correspond to heated areas mapped
   using measurements from the on-board temperature sensor.
   }\label{F:experiments_setup}
\end{figure}

We show our active sensing framework running in real-time
on a TurtleBot3 Waffle with an Intel Joule 570x
running Ubuntu Linux 16.04 and the Robot Operating System.
The experiments are conducted in a known indoor environment
using \ac{AMCL}\footnote{\url{wiki.ros.org/amcl}} receiving data from a LDS-01 laser distance scanner.
As shown in \Cref{SF:setup_environment},
for field mapping,
a temperature distribution in an empty $2.8$\,m\,$\times$\,$2.8$\,m area within the environment is generated
using a $2400$\,W radiant heater placed at one corner.
Measurements are taken using a
LM35 linear temperature sensor with a sensitivity of $10$\,mV/$\degree$C.

\begin{figure*}[!h]
\centering
  \begin{subfigure}{.51\columnwidth}
    \includegraphics[width=\columnwidth]{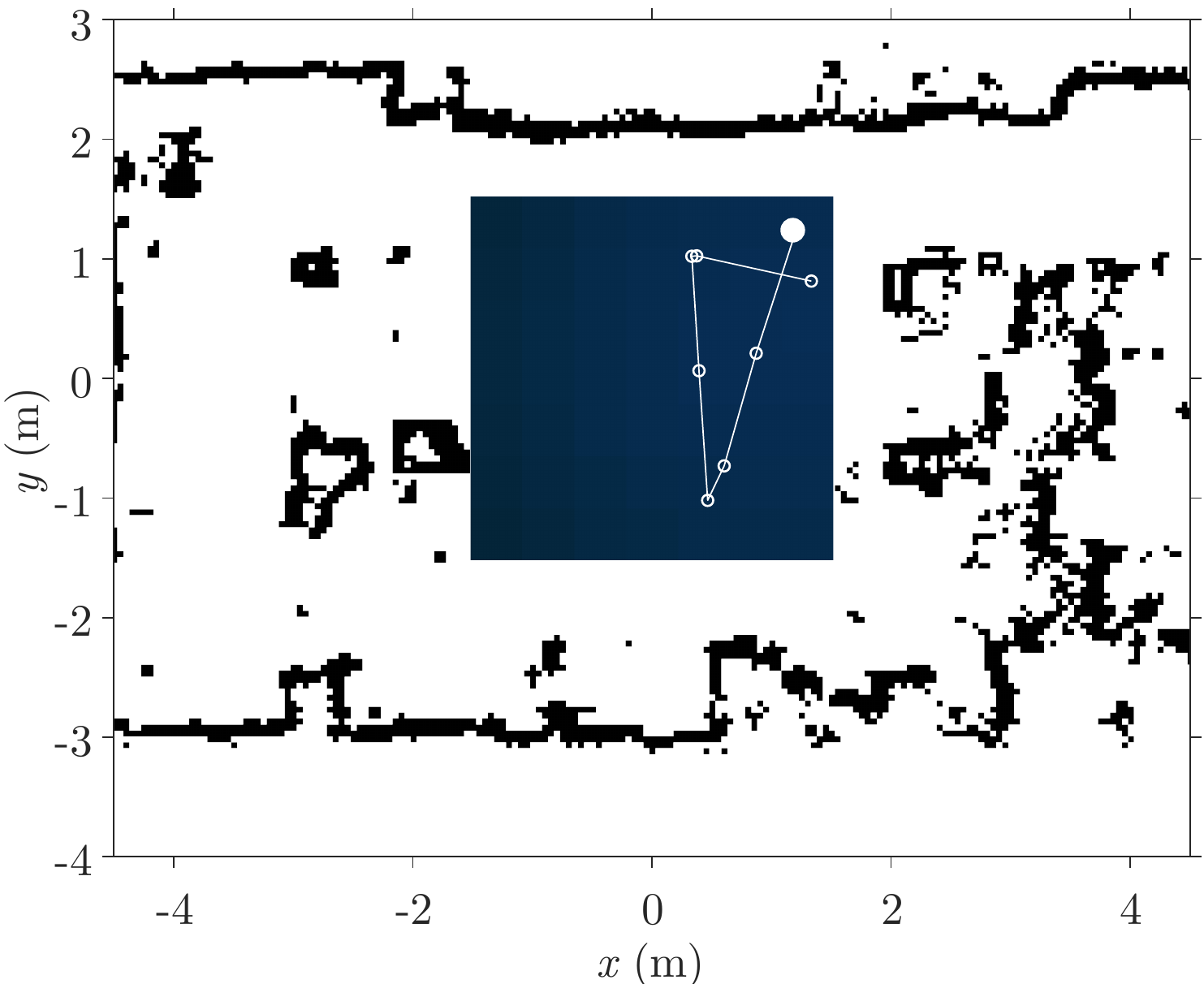}%
  %\caption{}%
  %\label{SF:map1}%
  \end{subfigure}\hfill%
  \begin{subfigure}{.51\columnwidth}
    \includegraphics[width=\columnwidth]{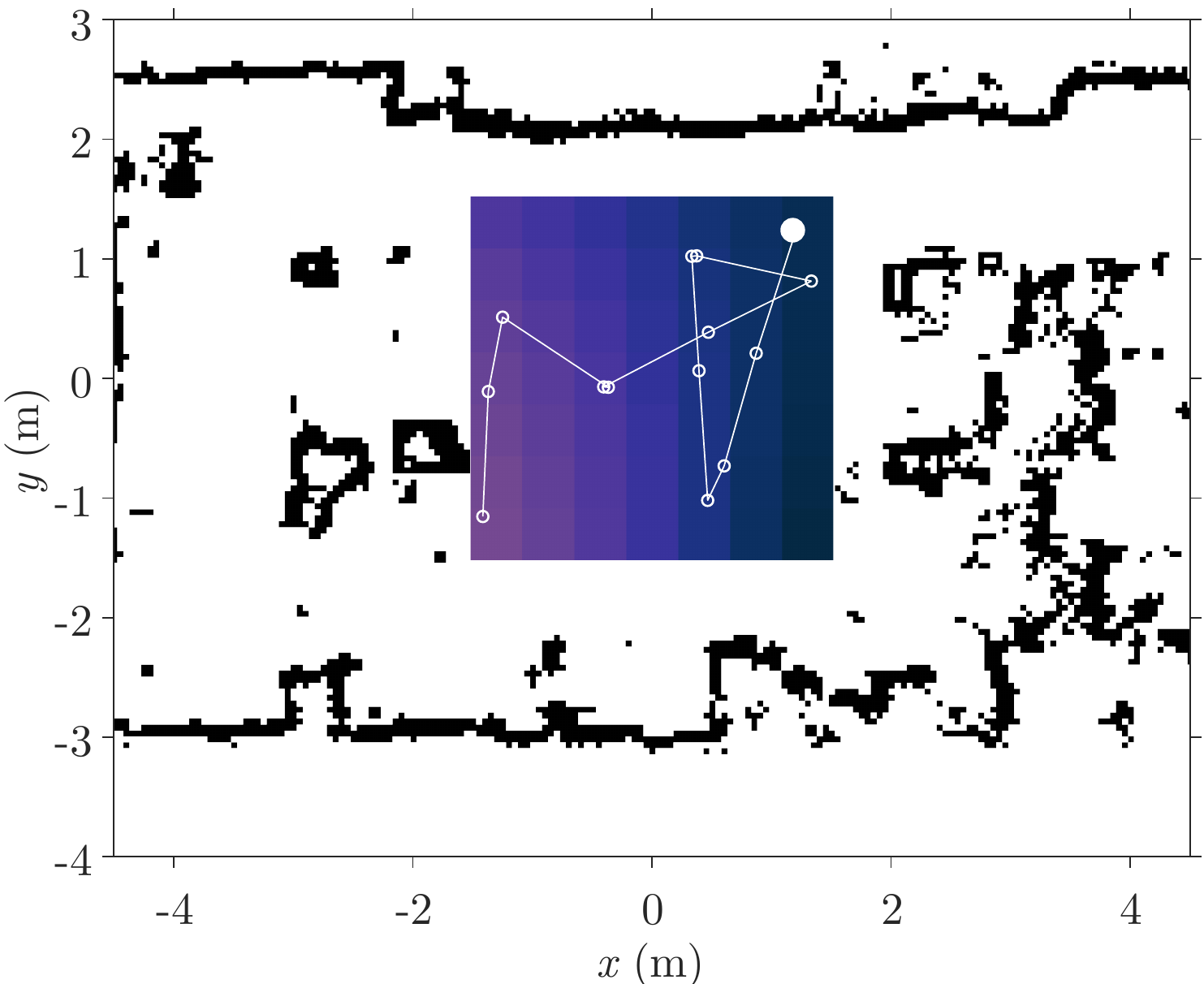}%
  %\caption{}%
  %\label{SF:map2}%
   \end{subfigure}\hfill%
  \begin{subfigure}{.51\columnwidth}
    \includegraphics[width=\columnwidth]{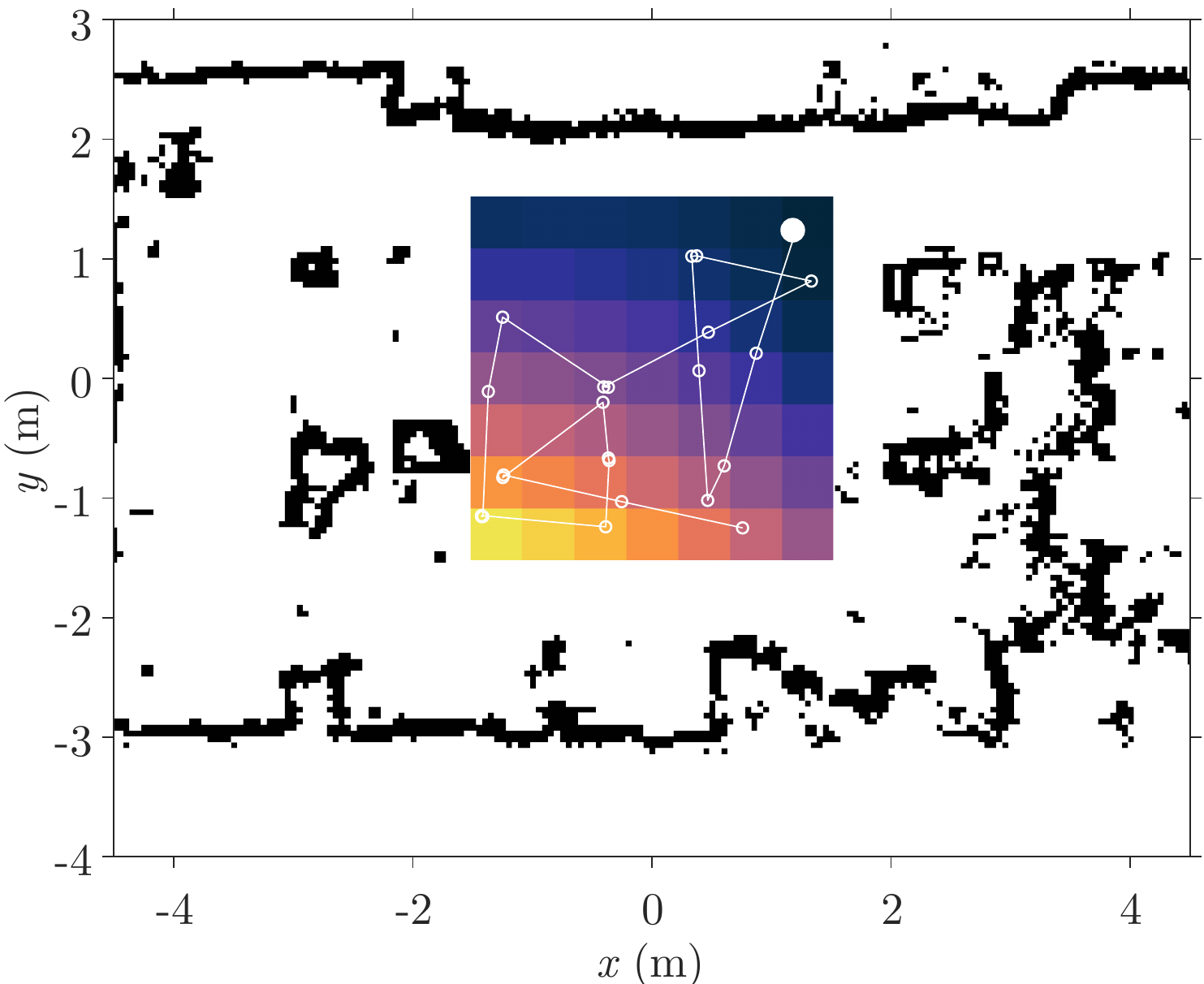}%
  %\caption{}%
  %\label{SF:map3}%
   \end{subfigure}
  \begin{subfigure}{.12\columnwidth}
    \raisebox{4mm}{\includegraphics[width=\columnwidth]{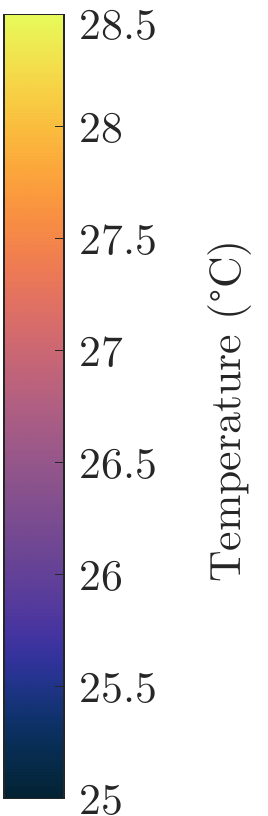}}%
  %\caption{}%
  %\label{SF:map3}%
   \end{subfigure} \hspace{1mm}
  \begin{subfigure}{.34\columnwidth}
    \includegraphics[width=\columnwidth]{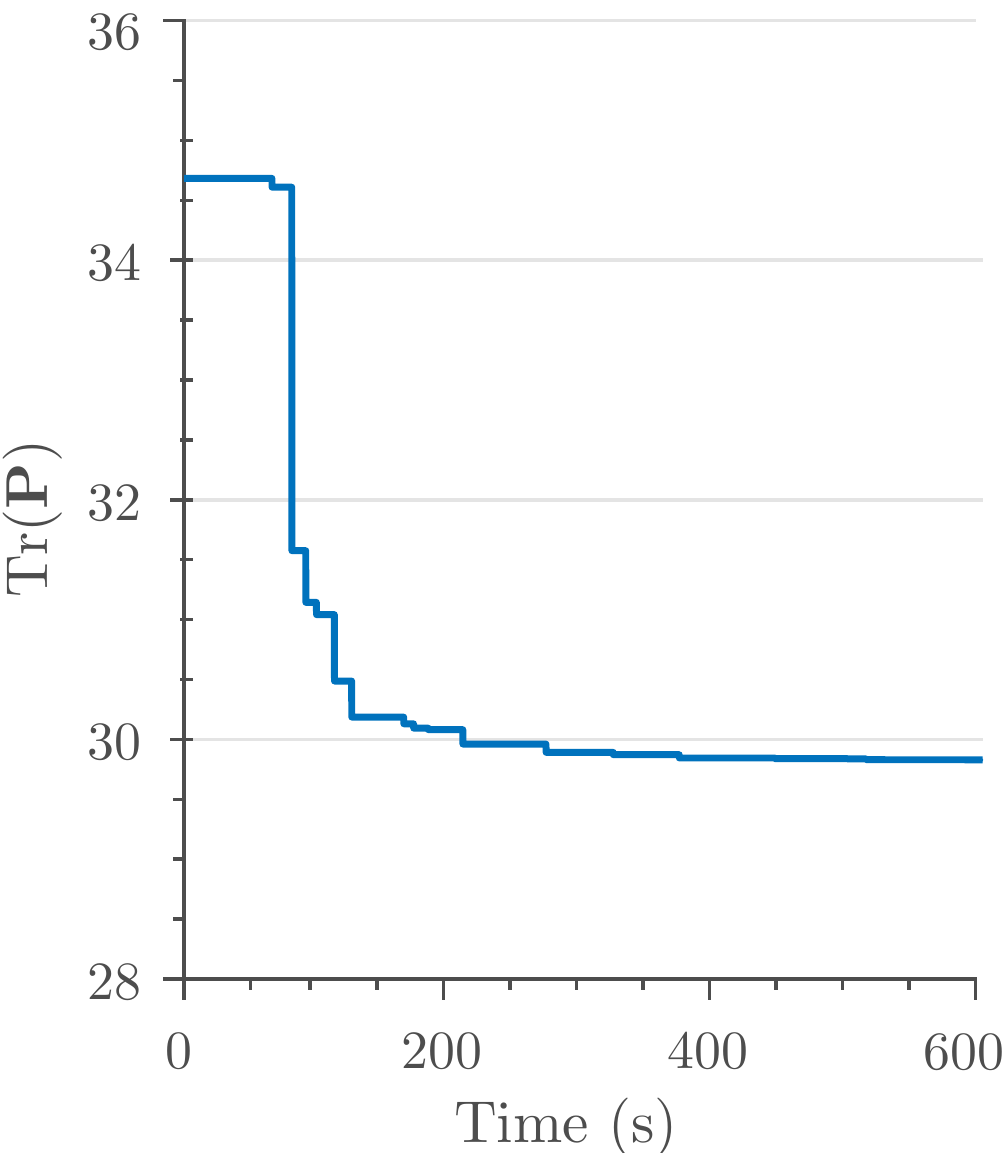}%
  %\caption{}%
  %\label{SF:uncertainty_plot}%
   \end{subfigure}%
   \caption{Experimental results of using our active sensing framework
   to map the indoor temperature distribution in \Cref{F:experiments_setup} in a $600$\,s mission.
   The three plots on the left depict the trajectories (white lines) and temperature field maps (colored gradients)
   at different snapshots of the mission at times $t = 100$\,s, $350$\,s, and $600$\,s.
   The white circles represent measurement sites,
   with the large solid one indicating the initial robot position.
   Yellower shades correspond to hotter regions.
   The sequence shows that our planner quickly explores the area,
   successfully detecting the heated corner (bottom-left) where the radiator is located.
   The curve on the right shows the uncertainty reduction in the field map over time,
   thus validating our approach.
   Note that planning time is taken into account.
   }\label{F:experiments_results}
\end{figure*}
A $0.4$\,m resolution grid is set for mapping with \acp{UI}.
To train the \ac{GP},
we follow the method in \Cref{SS:evaluation_mapping}
using manually gathered data within the target area
to obtain the hyperparameters% $\theta = \{m_{gp},\,\sigma_n^2,\,\sigma_f^2,\,l\} = \{23.64, \,3.6 \times 10^{-3}, \,1.21, \,5.23\}$
 for the \ac{SE} kernel.
As before,
the integral in \Cref{E:modified_kernel} is estimated using 5 Gauss-Hermite points.

Our uncertainty prediction method
is based on localization in a known environment using \ac{AMCL} (\Cref{SS:uncertainty_prediction}).
We sub-sample each candidate plan at $2$\,Hz
and estimate the robot pose
using a differential drive odometry model
and laser scans simulated in the known occupancy map.
For the odometry model,
variance parameters of $0.2$\,m$^2$ are used for Gaussian noise in both rotational and translational motion.

The aim is to show that our framework can map a realistic continuous field using a practical localization system.
The initial measurement point is ($0.2$\,m,\,$0.2$\,m) within the corner opposite the radiator.
We allocate a planning budget $B$ of $600$\,s.
Following the differential drive model,
each plan is piecewise linear
as defined by $N=3$ control waypoints
with a constant velocity of $0.26$\,m$/$s
and temperature measurements sampled at $0.25$\,Hz.
The planning objective is
our proposed R\'enyi-based utility function in \Cref{E:utility_renyi}.
Note that field map updates are triggered
upon allowing the sensor readings to stabilize
between successive measurement points.

\Cref{F:experiments_results} summarizes our experiments.
As expected,
the field map becomes more complete over time and uncertainty decreases
as the yellower heated region is discovered
in a successful proof of concept implementation.
%Note that the mean value of the \ac{GP}
%is lower than the range of measurements obtained
%due to the effects of heating and diffusion
%making it difficult to obtain a controlled temperature field
%as a reference.
%This motivates addressing temporal dynamics
%as a potential direction for future work.
\edit{Note that the values of the measurements obtained
are higher than that at which the field is initialized ($23.64\degree$),
i.e., the \ac{GP} mean.
This is due to the effects of heating and diffusion over time, after the training data was collected.
%which made it difficult to obtain a controlled temperature field in the room over time.
Future work will address these issues by investigating map representations that capture temporal, as well as spatial, dynamics.}

\section{CONCLUSIONS AND FUTURE WORK} \label{S:conclusions}
This work introduced an informative planning framework
for active mapping
that accounts for the robot pose uncertainty
in both the mapping and planning stages.
Our method uses \acp{GP} with \acp{UI} to propagate the robot pose uncertainty
into the model of a target environmental field.
For planning,
we introduced a new utility function
that tightly couples the uncertainties in the robot pose and field map
by applying the concept of R\'enyi's entropy in \ac{GP}-based mapping scenarios.
Our formulation enables the robot to trade off between exploration and exploitation
in a principled way, without relying on any manually tuned parameters.

Our framework was evaluated extensively in simulation.
We showed that it achieves more conservative exploratory behavior
compared to different planning and mapping strategies,
while producing more accurate maps.
Experimental validation was performed through a proof of concept deployment,
revealing its promise in future applications.

Future work will examine field models with temporal dynamics
and efficiency improvements %, e.g., by using submaps,
to handle more complex environments.
Other interesting research directions involve refining the uncertainty prediction method
and its relationship to the R\'{e}nyi parameter $\alpha$
for more reliable planning.

\bibliographystyle{IEEEtranN}
\footnotesize
\bibliography{references/2019-icra-popovic}

% Generated by IEEEtranN.bst, version: 1.14 (2015/08/26)
\begin{thebibliography}{24}
\providecommand{\natexlab}[1]{#1}
\providecommand{\url}[1]{#1}
\csname url@samestyle\endcsname
\providecommand{\newblock}{\relax}
\providecommand{\bibinfo}[2]{#2}
\providecommand{\BIBentrySTDinterwordspacing}{\spaceskip=0pt\relax}
\providecommand{\BIBentryALTinterwordstretchfactor}{4}
\providecommand{\BIBentryALTinterwordspacing}{\spaceskip=\fontdimen2\font plus
\BIBentryALTinterwordstretchfactor\fontdimen3\font minus
  \fontdimen4\font\relax}
\providecommand{\BIBforeignlanguage}[2]{{%
\expandafter\ifx\csname l@#1\endcsname\relax
\typeout{** WARNING: IEEEtranN.bst: No hyphenation pattern has been}%
\typeout{** loaded for the language `#1'. Using the pattern for}%
\typeout{** the default language instead.}%
\else
\language=\csname l@#1\endcsname
\fi
#2}}
\providecommand{\BIBdecl}{\relax}
\BIBdecl

\bibitem[Hitz et~al.(2017)Hitz, Galceran, Garneau, Pomerleau, and
  Siegwart]{Hitz2017}
G.~Hitz, E.~Galceran, M.-{\`{E}}. Garneau, F.~Pomerleau, and R.~Siegwart,
  ``{Adaptive Continuous-Space Informative Path Planning for Online
  Environmental Monitoring},'' \emph{Journal of Field Robotics}, vol.~34,
  no.~8, pp. 1427--1449, 2017.

\bibitem[Jadidi et~al.(2019)Jadidi, Miro, and Dissanayake]{Jadidi2019}
M.~G. Jadidi, J.~V. Miro, and G.~Dissanayake, ``Sampling-based incremental
  information gathering with applications to robotic exploration and
  environmental monitoring,'' \emph{The International Journal of Robotics
  Research}, vol.~38, no.~6, pp. 658--685, 2019.

\bibitem[Oliveira et~al.(2017)Oliveira, Ott, Guizilini, Ramos, and
  Sep]{Oliveira2017}
R.~Oliveira, L.~Ott, V.~Guizilini, F.~Ramos, and R.~O. Sep, ``{Bayesian
  Optimisation for Safe Navigation under Localisation Uncertainty},'' in
  \emph{International Symposium of Robotics Research}.\hskip 1em plus 0.5em
  minus 0.4em\relax Puerto Varas: Springer, 2017.

\bibitem[Carrillo et~al.(2018)Carrillo, Dames, Kumar, and
  Castellanos]{Carrillo2018}
H.~Carrillo, P.~Dames, V.~Kumar, and J.~A. Castellanos, ``{Autonomous robotic
  exploration using a utility function based on R{\'{e}}nyi's general theory of
  entropy},'' \emph{Autonomous Robots}, vol.~42, no.~2, pp. 235--256, 2018.

\bibitem[Popovi{\'{c}} et~al.(2019)Popovi{\'{c}}, Vidal-Calleja, Hitz, Chung,
  Sa, Siegwart, and Nieto]{Popovic2019}
M.~Popovi{\'{c}}, T.~Vidal-Calleja, G.~Hitz, J.~J. Chung, I.~Sa, R.~Siegwart,
  and J.~Nieto, ``{An informative path planning framework for UAV-based terrain
  monitoring},'' \emph{Autonomous Robots}, 2019, under review, arXiv preprint
  \href{http://arxiv.org/abs/1809.03870}{arXiv:1809.03870}.

\bibitem[Papachristos et~al.(2017)Papachristos, Khattak, and
  Alexis]{Papachristos2017}
C.~Papachristos, S.~Khattak, and K.~Alexis, ``{Uncertainty-aware Receding
  Horizon Exploration and Mapping using Aerial Robots},'' in \emph{IEEE
  International Conference on Robotics and Automation}.\hskip 1em plus 0.5em
  minus 0.4em\relax Singapore: IEEE, 2017, pp. 4568--4575.

\bibitem[Manfreda et~al.(2018)Manfreda, McCabe, Miller, Lucas, Madrigal,
  Mallinis, Dor, Helman, Estes, Ciraolo, M{\"{u}}llerov{\'{a}}, Tauro, de~Lima,
  de~Lima, Maltese, Frances, Caylor, Kohv, Perks, Ruiz-P{\'{e}}rez, Su, Vico,
  and Toth]{Manfreda2018}
S.~Manfreda, M.~F. McCabe, P.~E. Miller, R.~Lucas, V.~P. Madrigal, G.~Mallinis,
  E.~B. Dor, D.~Helman, L.~Estes, G.~Ciraolo, J.~M{\"{u}}llerov{\'{a}},
  F.~Tauro, M.~I. de~Lima, J.~L. de~Lima, A.~Maltese, F.~Frances, K.~Caylor,
  M.~Kohv, M.~Perks, G.~Ruiz-P{\'{e}}rez, Z.~Su, G.~Vico, and B.~Toth, ``{On
  the Use of Unmanned Aerial Systems for Environmental Monitoring},''
  \emph{Remote Sensing}, vol.~10, no.~4, 2018.

\bibitem[Hollinger and Sukhatme(2014)]{Hollinger2014}
G.~A. Hollinger and G.~S. Sukhatme, ``{Sampling-based robotic information
  gathering algorithms},'' \emph{International Journal of Robotics Research},
  vol.~33, no.~9, pp. 1271--1287, 2014.

\bibitem[Girard(2004)]{Girard2004}
A.~Girard, ``{Approximate methods for propagation of uncertainty with Gaussian
  process models},'' Ph.D. dissertation, University of Glasgow, 2004.

\bibitem[Mchutchon and Rasmussen(2011)]{Mchutchon2011}
A.~Mchutchon and C.~E. Rasmussen, ``{Gaussian Process Training with Input
  Noise},'' \emph{Advances in Neural Information Processing Systems}, pp.
  1341--1349, 2011.

\bibitem[Bourgault et~al.(2002)Bourgault, Makarenko, Williams, Grocholsky, and
  Durrant-Whyte]{Bourgault2002}
F.~Bourgault, A.~A. Makarenko, S.~B. Williams, B.~Grocholsky, and H.~F.
  Durrant-Whyte, ``{Information Based Adaptive Robotic Exploration},'' in
  \emph{IEEE/RSJ International Conference on Intelligent Robots and
  Systems}.\hskip 1em plus 0.5em minus 0.4em\relax Lauseanne: IEEE, 2002, pp.
  540--545.

\bibitem[Valencia et~al.(2012)Valencia, Mir{\'{o}}, Dissanayake, and
  Andrade-Cetto]{Valencia2012}
R.~Valencia, J.~V. Mir{\'{o}}, G.~Dissanayake, and J.~Andrade-Cetto, ``{Active
  Pose SLAM},'' in \emph{IEEE/RSJ International Conference on Intelligent
  Robots and Systems}, Vilamoura, 2012, pp. 1885--1891.

\bibitem[Whitehead and Ballard(2008)]{Whitehead2008}
S.~D. Whitehead and D.~H. Ballard, ``{Active Perception and Reinforcement
  Learning},'' \emph{Neural Computation}, vol.~2, no.~4, pp. 409--419, 2008.

\bibitem[Rasmussen and Williams(2006)]{Rasmussen2006}
C.~E. Rasmussen and C.~K.~I. Williams, \emph{{Gaussian Processes for Machine
  Learning}}.\hskip 1em plus 0.5em minus 0.4em\relax Cambridge, MA: MIT Press,
  2006.

\bibitem[Popovi{\'{c}} et~al.(2017{\natexlab{a}})Popovi{\'{c}}, Hitz, Nieto,
  Sa, Siegwart, and Galceran]{Popovic2017ICRA}
M.~Popovi{\'{c}}, G.~Hitz, J.~Nieto, I.~Sa, R.~Siegwart, and E.~Galceran,
  ``{Online Informative Path Planning for Active Classification Using UAVs},''
  in \emph{IEEE International Conference on Robotics and Automation}.\hskip 1em
  plus 0.5em minus 0.4em\relax Singapore: IEEE, 2017.

\bibitem[Popovi{\'{c}} et~al.(2017{\natexlab{b}})Popovi{\'{c}}, Vidal-Calleja,
  Hitz, Sa, Siegwart, and Nieto]{Popovic2017IROS}
M.~Popovi{\'{c}}, T.~Vidal-Calleja, G.~Hitz, I.~Sa, R.~Y. Siegwart, and
  J.~Nieto, ``{Multiresolution Mapping and Informative Path Planning for
  UAV-based Terrain Monitoring},'' in \emph{IEEE/RSJ International Conference
  on Intelligent Robots and Systems}.\hskip 1em plus 0.5em minus 0.4em\relax
  Vancouver: IEEE, 2017.

\bibitem[Kaelbling et~al.(1998)Kaelbling, Littman, and
  Cassandra]{Kaelbling1998}
L.~Kaelbling, M.~Littman, and A.~Cassandra, ``{Planning and Acting in Partially
  Observable Stochastic Domains},'' \emph{Artificial Intelligence}, vol. 101,
  no. 1-2, pp. 99--134, 1998.

\bibitem[Costante et~al.(2017)Costante, Delmerico, Werlberger, and
  Valigi]{Costante2017}
G.~Costante, J.~Delmerico, M.~Werlberger, and P.~Valigi, ``{Exploiting
  Photometric Information for Planning under Uncertainty},'' in \emph{Robotics
  Research}.\hskip 1em plus 0.5em minus 0.4em\relax Springer, 2017, pp.
  107--124.

\bibitem[Bry and Roy(2011)]{Bry2011}
A.~Bry and N.~Roy, ``{Rapidly-exploring random belief trees for motion planning
  under uncertainty},'' in \emph{IEEE International Conference on Robotics and
  Automation}.\hskip 1em plus 0.5em minus 0.4em\relax Shanghai: IEEE, 2011, pp.
  723--730.

\bibitem[Richter et~al.(2013)Richter, Bry, and Roy]{Richter2013}
C.~Richter, A.~Bry, and N.~Roy, ``{Polynomial Trajectory Planning for
  Aggressive Quadrotor Flight in Dense Indoor Environments},'' in
  \emph{International Symposium of Robotics Research}.\hskip 1em plus 0.5em
  minus 0.4em\relax Singapore: Springer, 2013.

\bibitem[Hansen(2006)]{Hansen2006}
N.~Hansen, ``{The CMA evolution strategy: A comparing review},'' \emph{Studies
  in Fuzziness and Soft Computing}, vol. 192, no. 2006, pp. 75--102, 2006.

\bibitem[Golshani and Pasha(2010)]{Golshani2010}
L.~Golshani and E.~Pasha, ``{R{\'{e}}nyi entropy rate for Gaussian
  processes},'' \emph{Information Sciences}, vol. 180, no.~8, pp. 1486--1491,
  2010.

\bibitem[Sol{\`{a}}(2017)]{Sola2017}
J.~Sol{\`{a}}, ``{Course on SLAM},'' Barcelona, 2017.

\bibitem[Kaess and Dellaert(2009)]{Kaess2009}
M.~Kaess and F.~Dellaert, ``{Covariance recovery from a square root information
  matrix for data association},'' \emph{Robotics and Autonomous Systems},
  vol.~57, no.~12, pp. 1198--1210, 2009.

\end{thebibliography}

\end{document}